\documentclass[times,twocolumn,final,authoryear]{elsarticle}

\usepackage{jasr}
\usepackage{framed,multirow}

\usepackage{amsmath, amssymb}
\usepackage{latexsym}

\usepackage{booktabs} 
\usepackage{graphicx}
\usepackage{bm}
\usepackage{makecell}
\usepackage{subcaption}

\usepackage[switch]{lineno}

\usepackage{url}
\usepackage{xcolor}
\definecolor{newcolor}{rgb}{.8,.349,.1}
\usepackage[citebordercolor=white]{hyperref}

\usepackage{soul}

\newcommand{\synthetic}{\texttt{synthetic}}
\newcommand{\lightbox}{\texttt{lightbox}}
\newcommand{\sunlamp}{\texttt{sunlamp}}
\newcommand{\prisma}{\texttt{prisma25}}

\newcommand{\Ehead}{h_\text{E}}
\newcommand{\Hhead}{h_\text{H}}
\newcommand{\Shead}{h_\text{S}}

\journal{Advances in Space Research}

\begin{document}

\verso{Tae Ha Park \textit{etal}}

\begin{frontmatter}

\title{Robust multi-task learning and online refinement for spacecraft pose estimation across domain gap}

\author[1]{Tae Ha \snm{Park}\corref{cor1}}
\cortext[cor1]{Corresponding author}
\ead{tpark94@stanford.edu}

\author[1]{Simone D'Amico}
\ead{damicos@stanford.edu}

\address[1]{Department of Aeronautics \& Astronautics, Stanford University, 496 Lomita Mall, Stanford, CA 94305, USA}


\begin{abstract}
This work presents Spacecraft Pose Network v2 (SPNv2), a Convolutional Neural Network (CNN) for pose estimation of noncooperative spacecraft across domain gap. SPNv2 is a multi-scale, multi-task CNN which consists of a shared multi-scale feature encoder and multiple prediction heads that perform different tasks on a shared feature output. These tasks are all related to detection and pose estimation of a target spacecraft from an image, such as prediction of pre-defined satellite keypoints, direct pose regression, and binary segmentation of the satellite foreground. It is shown that by jointly training on different yet related tasks with extensive data augmentations on synthetic images only, the shared encoder learns features that are common across image domains that have fundamentally different visual characteristics compared to synthetic images. This work also introduces Online Domain Refinement (ODR) which refines the parameters of the normalization layers of SPNv2 on the target domain images online at deployment. Specifically, ODR performs self-supervised entropy minimization of the predicted satellite foreground, thereby improving the CNN's performance on the target domain images without their pose labels and with minimal computational efforts. The GitHub repository for SPNv2 is available at \url{https://github.com/tpark94/spnv2}.
\end{abstract}


\end{frontmatter}


\section{Introduction}
Monocular-based pose estimation of a noncooperative spacecraft has been a topic of research interest in recent years due to its applicability to various future mission concepts that address sustainability of the near-Earth space, such as refueling space assets \citep{Reed2016RestoreL} and active debris removal \citep{Forshaw2016Removedebris}. In such scenarios, an autonomous servicer spacecraft could generate safe and fuel-efficient rendezvous and docking trajectories based on real-time estimated pose of the target relative to the servicer. The pose information can be extracted from a sequence of images captured by the monocular camera, a low Size-Weight-Power-Cost (SWaP-C) sensor that is particularly suitable to the limited on-board capacity of small satellites such as CubeSats.

In recent years, Machine Learning (ML) techniques based on Convolutional Neural Networks (CNN) have been extensively studied for the application of spaceborne computer vision and especially pose estimation \citep{Park2019AAS, PasqualettoCassinis2021Coupled, Black2021PoseEstimationCygnus, Chen2019SatellitePE}, especially with the advent of the Spacecraft PosE Estimation Dataset (SPEED) \citep{Sharma2019AAS, Sharma2019SPEEDonSDR} and the international Satellite Pose Estimation Competition (SPEC2019) \citep{Kisantal2020SPEC} co-organized by the Advanced Concepts Team (ACT) of the European Space Agency (ESA) and the Space Rendezvous Laboratory (SLAB) at Stanford University. SPEED consists of 15,000 synthetic images of the Tango spacecraft from the PRISMA mission \citep{PRISMA_chapter, Damico2014IJSSE} and 300 Hardware-In-the-Loop (HIL) images captured from the Testbed for Rendezvous and Optical Navigation (TRON) facility at SLAB \citep{Park2021AAS}. It allows the community to benchmark pose estimation performance and compare different methods on a unified set of test data with consistent metrics. However, spaceborne images have fundamentally different visual properties compared to computer-generated synthetic images, so CNN performance on synthetic images does not translate to the equal level of performance on spaceborne images encountered during a mission. Known as domain gap in literature \citep{BenDavid2010LearningFromDiffDomains, Peng2017VisDA}, addressing the performance gap between training and testing data from different distributions is an active field of research in a wide range of deep learning applications. However, addressing domain gap in spaceborne applications is extremely difficult compared to terrestrial applications such as autonomous driving, since access to space is prohibitively expensive for data collection and road tests. While SPEED's HIL images are designed as surrogates of spaceborne images, they lack in quantity and diversity in pose labels and illumination conditions to comprehensively evaluate the generalization capability of the CNN models across different image domains.

To address this issue, SPEED+ \citep{Park2021speedplusSDR, Park2021speedplus} has been recently introduced with two distinct HIL image domains, $\lightbox$ and $\sunlamp$, that are captured from the upgraded TRON facility at SLAB \citep{Park2021AAS}. The $\lightbox$ domain consists of 6,740 images of the half-scale mockup model of the Tango spacecraft illuminated with albedo lightboxes to simulate diffuse light in the Earth's orbit, whereas the $\sunlamp$ domain contains 2,791 images of the same model illuminated with a metal halide arc lamp to simulate direct sunlight. With augmented realism of the visual characteristics of the HIL images, SPEED+ allows for a comprehensive analysis of the robustness of the CNN models trained exclusively on synthetic images. The SPEED+ dataset was used in the second international Satellite Pose Estimation Competition (SPEC2021) with emphasis on bridging the performance gap between the synthetic training and HIL test images \citep{park2023spec2021}.

This paper introduces Spacecraft Pose Network v2 (SPNv2), the SLAB's solution to resolving the domain gap in SPEED+ and beyond. Similar to its predecessor \citep{Sharma2019AAS, Sharma2020TAES} which consists of the shared AlexNet \citep{Krizhevsky2012AlexNet} backbone and multiple heads for bounding box detection, attitude classification and residual regression, SPNv2 is a multi-scale, multi-task pose estimation CNN with a shared feature encoder and prediction heads that perform various tasks such as bounding box detection, pose regression, heatmap prediction and spacecraft foreground segmentation. These tasks are different yet rely on common information such as the target's geometry and pose. Therefore, jointly training SPNv2 on those tasks forces the shared encoder to learn such common information, which also translates across different image domains, instead of any task-specific features that may also be particular to the synthetic image domain.

Training SPNv2 consists of two stages: 
\begin{enumerate}
	\item \textbf{offline robust training}, which trains exclusively on the SPEED+ $\synthetic$ training set with extensive data augmentation to make CNN as domain-invariant as possible, and
	\item \textbf{online domain refinement}, which refines the feature encoder of SPNv2 on unlabeled test images by minimizing the Shannon entropy \citep{Shannon1948} on the segmentation task.
\end{enumerate}
The motivation of the Online Domain Refinement (ODR) stage is that, despite the realism of SPEED+ HIL images, there still remains an inherent gap between the $\lightbox$, $\sunlamp$ and spaceborne domains. Naturally, while good performances on SPEED+ HIL domains suggest good generalization capability of the CNN models, it does not necessarily translate to an equal level of performance on the spaceborne images. Therefore, the second stage aims to directly incorporate the test or target domain images into the training procedure, so that the CNN model parameters are influenced by the exact images on which it is designed to function. While this paper performs ODR on SPEED+ test domains, the same strategy can be applied to spaceborne images acquired during missions. Note that ODR is \emph{source-free}, i.e., it does not require simultaneous access to the large-scale training data for adaptation, which complies with the memory and computational constraints of the on-board avionics.

It is shown that the offline training of SPNv2 alone provides remarkable performance on SPEED+ HIL domains compared to the baseline studies \citep{Park2021speedplus}, with translation errors around 22cm / 23cm and orientation errors about 8.0$^\circ$ / 10.4$^\circ$ for $\lightbox$ and $\sunlamp$, respectively, for a larger, batch-agnostic variant of SPNv2. ODR further refines the translation error by another 6 cm / 7 cm and orientation error by 2.4$^\circ$ / 0.6$^\circ$ on $\lightbox$ and $\sunlamp$, respectively, after observing 4096 unlabeled images. Extensive analyses are performed on both offline training and ODR to articulate the factors contributing to this success.

This paper is organized as follows. Section 2 describes related state-of-the-art for different components of the proposed methods. Then, Sections 3 and 4 explain the proposed SPNv2 architecture, offline training procedure, and ODR. Section 5 examines the experimental results of both offline training and ODR, and Section 6 discusses limitations and future directions. The paper ends with concluding remarks in Section 7.

\begin{figure*}[!t]
	\centering
	\includegraphics[width=1.0\textwidth]{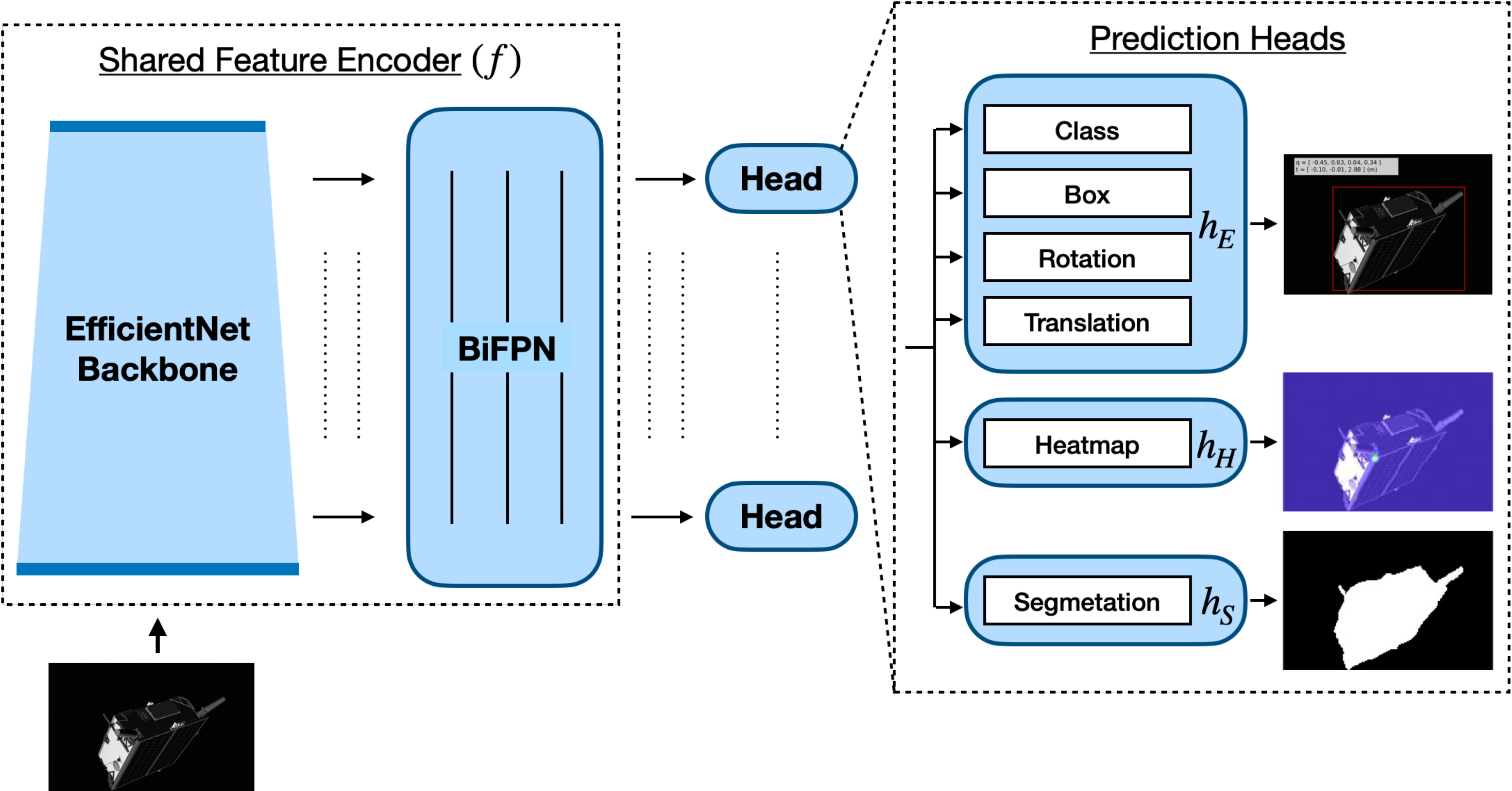}
	\caption{The pose estimation CNN architecture based on the EfficientNet \citep{Tan2019EfficientNetICML} backbone, the Bi-directional Feature Pyramid Network (BiFPN), and multi-task head networks shared by the features at all scales.}
	\label{fig:architecture}
\end{figure*}

\section{Related Work}

\paragraph{Spacecraft Pose Estimation}
The first ML-based approach to pose estimation of a known target spacecraft is Spacecraft Pose Network (SPN) \citep{Sharma2019AAS, Sharma2020TAES}, which performs relative attitude determination via a hybrid approach of attitude classification and regression and performs translation estimation by exploiting the perspective transformation and geometric constraints. Afterwards, many top-performing entries of SPEC2019 have shown diverse CNN architectures and pose estimation strategies, such as probablistic orientation estimation via soft classification \citep{Proenca2019Photorealistic} or estimation of a set of designated keypoints on the spacecraft surface via direction regression \citep{Park2019AAS} or heatmaps \citep{Chen2019SatellitePE}. Specifically, the estimated 2D keypoints can then be used in Perspective-$n$-Point (P$n$P) \citep{Lepetit2008EPnP} to recover full 6D pose. Many CNN models that followed the competition also adopt similar strategies as well \citep{Black2021PoseEstimationCygnus, PasqualettoCassinis2021Coupled}. Due to the limitation of SPEED and other proposed datasets for spacecraft pose estimation, these CNN models are designed to drive performance on synthetic images that have vastly different visual characteristics compared to spaceborne images. The readers are referred to \citet{PasqualettoCassinis2019Survey} for a more comprehensive review of monocular spacecraft pose estimation using both conventional and deep learning-based methods.

\paragraph{Domain Gap in Space}
Domain gap \citep{BenDavid2010LearningFromDiffDomains, Peng2017VisDA, Ganin2015DANN_ICML} is a ubiquitous problem for any real-life applications of ML models. However, addressing domain gap in spaceborne vision applications is extremely challenging due to the inaccessibility of space for even a small-scale data collection. This barrier prevents a comprehensive evaluation of the trained model on the spaceborne images on which the model should demonstrate robust performance \emph{prior} to deployment. Therefore, existing works have performed evaluation on a handful of spaceborne images from the previous missions \citep{Park2019AAS, Sharma2020TAES, Proenca2019Photorealistic, Black2021PoseEstimationCygnus} or captured in a high-fidelity simulation environment with a physical model of the target satellite \citep{Park2021speedplus, Kisantal2020SPEC, PasqualettoCassinis2021ORGL}. In these cases, the CNN models are trained on abundant synthetic images with extensive data augmentation or domain randomization \citep{Tobin2017DomainRandomization} to render them as domain-agnostic as possible. However, their evaluation on spaceborne or other physical image domains lacks comprehensiveness due to limited sample quantity, and they often lack in-depth analyses on the factors that attribute to successfully bridging the domain gap.

\paragraph{Source-Free Domain Adaptation}
Another consequence of inaccessibility of space is that the solution precludes many conventional domain adaptation algorithms \citep{Ganin2015DANN_ICML, Tzeng2017CVPR_ADDA, Sun2016ECCV_DeepCORAL} that assume simultaneous availability of both labeled source and unlabeled target domain data. In reality, the spaceborne images are only available during missions in space, thus performing domain adaptation on-board the spacecraft with loaded source data is simply not feasible given the computational and memory limitations of satellite avionics. Existing \emph{source-free} domain adaptation methods would leverage generative models during offline training \citep{Li2020ECCVModelAdaptationUDA, Kundu2020CVPRUniversalSourceFree}, pseudo-labeling and information maximization \citep{Liang2020ICMLSHOT}, entropy minimization \citep{Wang2021ICLRTENT}, and so on. Test-Time Training (TTT) \citep{Sun19TestTimeTraining, Liu2021TTT++} jointly optimizes the main task (e.g., classification) and a Self-Supervised Learning (SSL) task from a shared feature encoder. Then, at test time, the encoder parameters are updated via self-supervised learning on the target domain samples. While it is a new paradigm to source-free domain adaptation, TTT requires an additional head for SSL that is often hand-designed by users (e.g., rotation prediction \citep{Gidaris2018RotationPrediction}, jigsaw puzzle \citep{Noroozi2016ECCVJigsawPuzzle}) or requires a large batch of negative sample pairs as is the case for contrastive learning \citep{Chen2020ICMLContrastiveLearning}. Instead, this work minimizes entropy of the predicted satellite foreground and localizes the gradient update to the affine transformation parameters of the batch normalization layers \citep{Wang2021ICLRTENT, Li2016AdaBN}, which does not require an additional self-supervised proxy task and minimizes the computation incurred during backpropagation.

\section{Offline Robust Training}

This section describes the offline robust training of SPNv2 on the synthetic dataset, which is achieved by combination of a multi-scale, multi-task CNN architecture design, extensive data augmentation and domain randomization.

\subsection{Multi-Scale, Multi-Task Architecture}

The main SPNv2 architecture visualized in Figure \ref{fig:architecture} closely follows EfficientPose \citep{Bukschat2020EfficientPose} based on the EfficientDet \citep{Tan2020EfficientDetCVPR} feature encoder, which comprises the EfficientNet \citep{Tan2019EfficientNetICML} backbone and Bi-directional Feature Pyramid Network (BiFPN) to fuse features from different scales. Then, the output of the shared feature encoder is fed into different prediction heads. Following the original design, the EfficientPose ($\Ehead$) head consists of the subnets for binary classification of the object presence, bounding box prediction, target rotation and translation regression. In this work, two additional prediction heads are added: the Heatmap ($\Hhead$) head which predicts the 2D heatmaps associated with each pre-designated keypoints on the spacecraft surface, and the Segmentation ($\Shead$) head for pixel-wise binary segmentation of the spacecraft foreground. The motivation for the proposed architecture is that the multi-task learning \citep{Caruana1997MultitaskLearning} of related tasks could improve the generalization capability by suppressing the shared encoder from learning any task-specific features. By jointly training on related yet different tasks, the encoder would instead learn features that are common to those tasks, such as the spacecraft geometry and pose. Note that these features are also common across different image domains; therefore, learning them would also improve generalization from synthetic training to spaceborne test images. Henceforth, SPNv2 with $\Hhead$, $\Ehead$, and $\Shead$ is referred to be in full configuration.

Similar to EfficientPose \citep{Bukschat2020EfficientPose}, the convolutional parameters in the prediction heads are shared across different scales. The layouts of both the BiFPN feature fusion network and individual heads depend on the EfficientNet scaling parameter $\phi$ \citep{Tan2019EfficientNetICML}. The rotation and translation networks in $\Ehead$ also include the iterative refinement subnets whose layouts depend on $\phi$ as designed in the original work \citep{Bukschat2020EfficientPose}. The inputs to $\Ehead$ are the level 3-7 feature outputs from BiFPN, where level $n$ feature has the resolution downscaled from the input image by a factor of $2^n$. The subnets of $\Hhead$ and $\Shead$ mimic the structure of the bounding box subnet except the output dimension. However, in contrast to $\Ehead$, these heads instead use high-resolution level 2 feature output for predictions, following the practices of the original EfficientDet \citep{Tan2020EfficientDetCVPR} for semantic segmentation. Therefore, BiFPN is extended to always fuse the level 2-7 features, and the intermediate layers of these heads have the dimension $\min\{2W_\text{bifpn}(\phi), 256\}$, where $W_\text{bifpn}$ is the width of the BiFPN outputs at each level. The readers are referred to \citet{Tan2020EfficientDetCVPR} and \citet{Bukschat2020EfficientPose} for details on different feature levels and the scaling parameter $\phi$.

\subsection{Normalization Layers}

\begin{figure}[!t]
	\centering
	\includegraphics[width=0.49\textwidth, trim=4 4 4 4,clip]{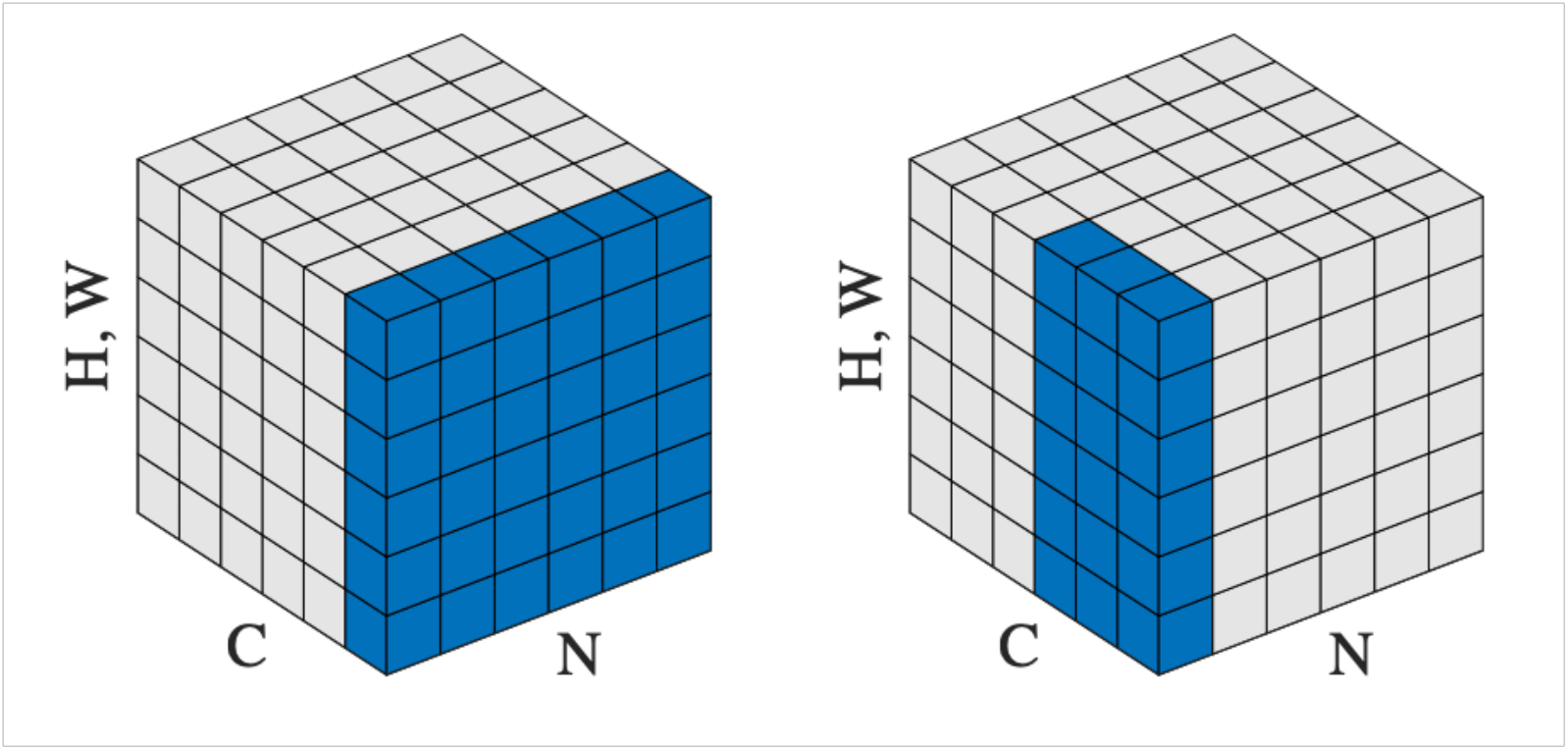}
	\caption{Visualization of a 4D feature map tensor across $N$ batches, $C$ channels, $H \times W$ spatial resolution in BN (\emph{left}) and GN (\emph{right}) layers. The pixels in blue are normalized by the same mean and variance. Diagram from \citet{Wu2018ECCV_GroupNorm}.}
	\label{fig:normalization layers}
\end{figure}

\begin{table*}[!t]
	\caption{List of data augmentations and equivalent commands in Albumentations \citep{Buslaev2020Albumentations}. Each augmentations are activated with 50\% probability.}
	\label{tab:data augmentations}
	\centering
	\tabcolsep=0.1cm
	\begin{tabular}{@{}cc@{}}
		\toprule
		Augmentation & Commands \\
		\midrule
		Brightness \& Contrast & \texttt{RandomBrightnessContrast} \\
		Random Erase \citep{Zhong2020RandomErasing} & \texttt{CoarseDropout} \\
		Sun Flare & \texttt{RandomSunFlare} \\
		Blur & \texttt{OneOf}(\texttt{MotionBlur}, \texttt{MedianBlur}, \texttt{GlassBlur}) \\
		Noise & \texttt{OneOf}(\texttt{GaussNoise}, \texttt{ISONoise}) \\
		\bottomrule
	\end{tabular}
\end{table*}

\begin{figure*}[!t]
	\centering
	\includegraphics[width=\textwidth,trim=4 4 4 4,clip]{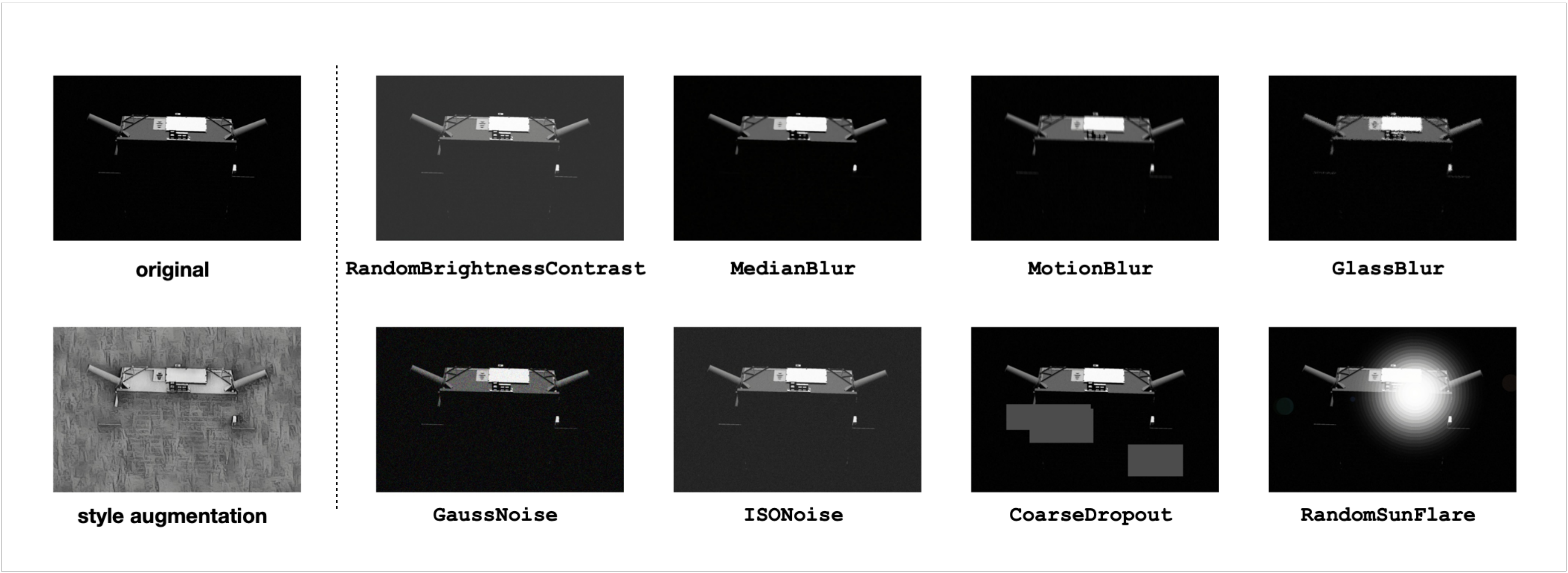}
	\caption{Example image, its stylized version, and visualization of different augmentation techniques.}
	\label{fig:augmentation visualize}
\end{figure*}

The original implementation of EfficientPose \citep{Bukschat2020EfficientPose} utilizes Group Normalization (GN) layers \citep{Wu2018ECCV_GroupNorm} throughout the entire network, which is a batch-agnostic alternative to the commonly employed Batch Normalization (BN) layers \citep{Ioffe2015BatchNorm}. As shown in Figure \ref{fig:normalization layers}, BN normalizes the features across $N$ images with the approximate mean and variance of the features across all images in the image domain, which are estimated using running average of the batch-wise feature statistics during training. On the other hand, GN normalizes the feature across a group of $G$ channels per each image using the statistics of the incoming features. Since the normalization procedure does not stretch across the batch of images, GN is a completely batch-agnostic layer. \citet{Wu2018ECCV_GroupNorm} have demonstrated that a CNN with GN layers can show equivalent or better performance on a number of benchmark computer vision tasks than a CNN with BN layer. The independence from batch-wise information means a CNN with GN layers can be trained on single images instead of batches in a training environment with limited computational capabilities. Moreover, as explained in Section \ref{section:ODR} and shown in Section \ref{section:Experiments} later, the batch-agnostic nature of GN layers favors the online refinement procedure against the new image domain. 

In this work, the prediction heads of SPNv2 use GN layers with 16 channels per group following the implementation of EfficientPose \citep{Bukschat2020EfficientPose}. However, for the EfficientNet backbone and the BiFPN layers, both BN layers and GN layers with the group size 8 are implemented and evaluated for its contribution to resolving the domain gap.

\subsection{Training Losses}

Training $\Ehead$ partially follows the original implementations of EfficientPose \citep{Bukschat2020EfficientPose} with a total of 9 anchor boxes of aspect ratios \{0.5, 1, 2\} at scales \{$2^{1/3}, 2^{2/3}, 2$\}. The classification subnet employs the focal loss \citep{Lin2017FocalLoss} with $\alpha = 0.25$ and $\gamma = 2.0$. Unlike EfficientPose, the bounding box subnet training minimizes the complete Intersection-over-Union (IoU) loss \citep{Zheng2020DIoU}, and the rotation and translation prediction subnets jointly minimize the pose error ($E_\text{pose}$) \citep{Kisantal2020SPEC}, also known as SPEED score, defined as
\begin{align}
	E_\text{pose} = E_\text{R}(\tilde{\bm{R}}, \bm{R}) + E_\text{T}(\tilde{\bm{t}}, \bm{t}) / \|\bm{t}\|, \label{eqn:speed score}
\end{align}
where, given the predicted and ground-truth rotation matrices and translation vectors $(\tilde{\bm{R}}, \tilde{\bm{t}})$ and $(\bm{R}, \bm{t})$, the rotation error $E_\text{R}$ and the translation error $E_\text{T}$ are respectively defined as
\begin{align}
	E_\text{R}(\tilde{\bm{R}}, \bm{R}) &= \arccos \frac{\text{tr}(\bm{R}^\top \tilde{\bm{R}}) - 1}{2}, \label{eqn:rotation error} \\
	E_\text{T}(\tilde{{\bm{t}}}, \bm{t}) &= \| \tilde{\bm{t}} - \bm{t} \|. \label{eqn:translation error}
\end{align}
The SPEED loss is then the sum of the rotation error in radians and translation error normalized by the ground-truth distance. Since SPEED loss is the official performance metric of the Satellite Pose Estimation Competition (SPEC), it is a natural choice of loss for the pose regression tasks. Note that while the EfficientPose implementation regresses the axis-angle representation of the orientation, the rotation subnet in this work instead regresses the 6D representation studied by \citet{Zhou2019RotationParametrization}, whose continuous property has shown to outperform other parametrization methods in rotation regression tasks. With the 6D representation as a pair of 3D vectors ($\bm{r}_1, \bm{r}_2$), the rotation matrix $\bm{R} = [\bm{R}_1 ~|~ \bm{R}_2 ~|~ \bm{R}_3 ]$ can be recovered column-wise via
\begin{align}
	\begin{cases}
		\bm{R}_1 &= N(\bm{r}_1) \\
		\bm{R}_2 &= N(\bm{r}_2 - (\bm{r}_1 \cdot \bm{r}_2) \bm{r}_1) \\
		\bm{R}_3 &= \bm{R}_1 \times \bm{R}_2
	\end{cases}
\end{align}
where $N(\cdot)$ is a normalization operator.

Finally, $\Hhead$ minimizes the pixel-wise mean squared error loss against the ground-truth heatmaps centered around each visible keypoints. $\Shead$ minimizes the pixel-wise binary cross entropy loss. All losses are optimized simultaneously with equal weights.

\begin{figure*}[!t]
	\centering
	\includegraphics[width=1.0\textwidth, trim=5 5 5 5,clip]{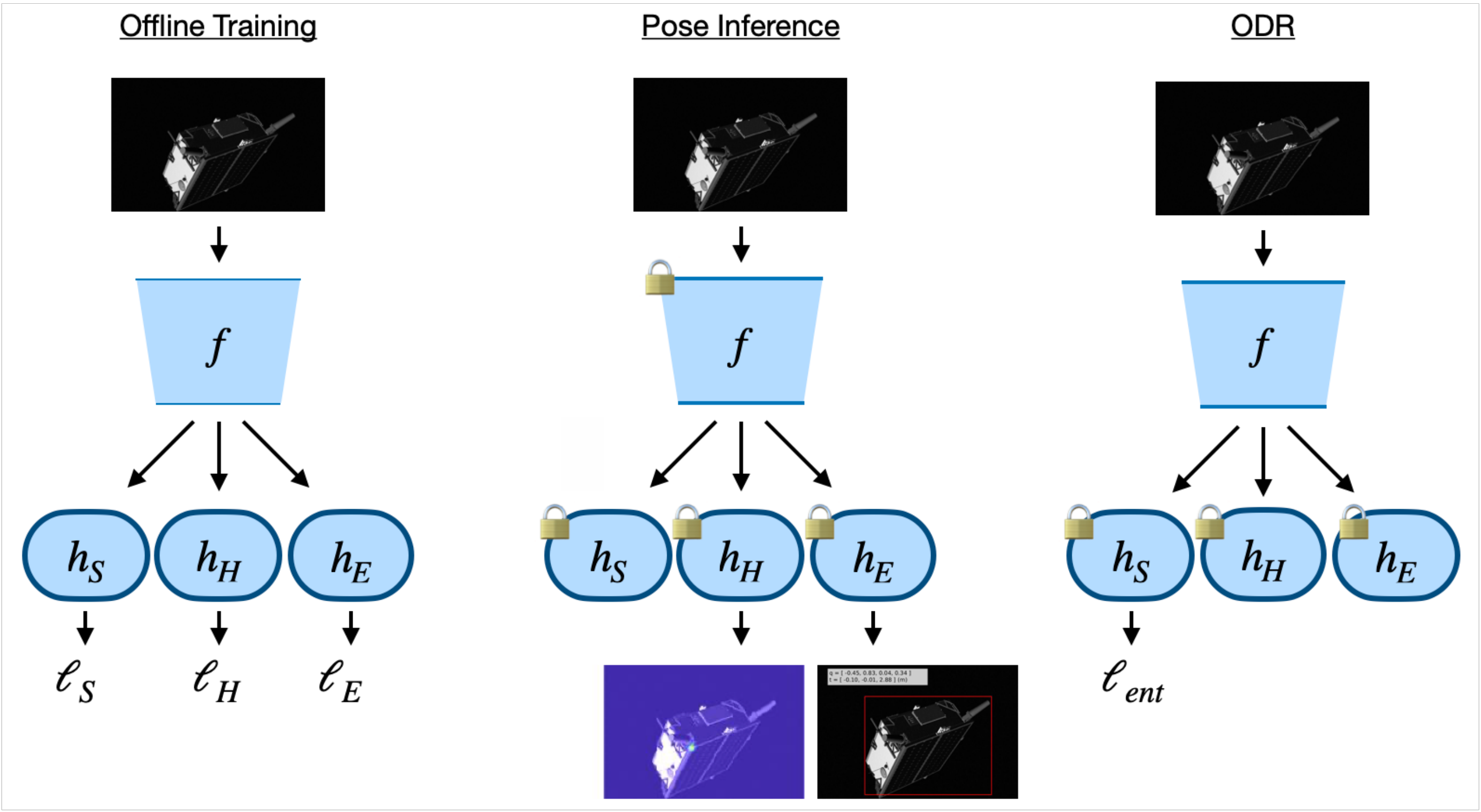}
	\caption{Visualization of complete workflow of offline training, pose inference, and ODR. The lock diagram indicates that the parameters of the corresponding component are fixed.}
	\label{fig:flowchart}
\end{figure*}

\subsection{Data Augmentation}

SPNv2 is trained with extensive data augmentation \citep{Shorten2019DataAugmentation} to mitigate the overfitting to the synthetic images. The augmentations are implemented with the Albumentations library \citep{Buslaev2020Albumentations}, and the list of employed augmentations and their equivalent Albumentations commands are provided in Table \ref{tab:data augmentations}. Each of the augmentations are activated with probability 0.5. Note that when applying blur, either motion blur, median blur, or Gaussian blur is applied with equal probability, and likewise for applying noise as well. Moreover, random erase \citep{Zhong2020RandomErasing} and sun flare augentations, which simulate the occlusion of satellite parts due to extreme shadowing or the sun lamp's direct sunlight, are implemented after modification to localize the effect within the bounding box of the target satellite instead of the entire image frame.

In addition to the standard data augmentation techniques, style augmentation \citep{Jackson2019ICCV_StyleAug} is employed to randomize the style and inherent texture of the synthetic spacecraft via neural style transfer \citep{Ghiasi2017StyleTransfer}, as done in the SPEED+ baseline studies \citep{Park2021speedplus}. An example training image, its style-augmented version, and employed augmentation techniques are visualized in Figure \ref{fig:augmentation visualize}.

\section{Online Domain Refinement} \label{section:ODR}

While the offline robust training on synthetic images could improve the generalizability across different spacecraft image domains, its performance during real missions lacks fundamental guarantee as the spaceborne images have never been integrated into the offline training procedure. Therefore, as the spaceborne images of the target become available during the rendezvous phase, SPNv2 can be refined on the target domain images on-board the spacecraft avionics. Considering the on-board hardware limitations, there exists a number of constraints that must be met for a fully mission-compliant ODR method; specifically, it must be 
\begin{enumerate}
	\item \emph{source-free}, i.e., the satellite only has access to a model trained offline on synthetic data;
	\item \emph{sequential}, i.e., it must not wait to collect a large batch of samples to perform refinement while withholding predictions;
	\item \emph{computationally efficient}, i.e., any learning components must be minimized or restricted to a small subset of the entire CNN.
\end{enumerate}
This paper adopts and modifies TENT \citep{Wang2021ICLRTENT} to minimize Shannon entropy \citep{Shannon1948} on the pixel-wise binary classification of the segmentation head. Specifically, the gradient updates are restricted to the affine transformation parameters of the normalization layers of the shared feature encoder of SPNv2. If the encoder contains the BN layers, the ODR must re-estimate the batch-wise feature normalization statistics as well. Unlike TENT which normalizes the features using the input batch statistics, the proposed method continuously updates the running statistics of the BN layers after every $B$ images in a manner similar to AdaBN \citep{Li2016AdaBN}. If the encoder instead consists of the GN layers, then the process of estimating the batch-wise statistics can be omitted altogether. The proposed ODR is also similar to TTT \citep{Sun19TestTimeTraining, Liu2021TTT++} in that both works are essentially multi-task learning, and the network parameters are updated only in the shared feature encoder. However, unlike TTT which updates the entire encoder parameters, ODR only updates the affine transformation parameters of the normalization layers that are only a small fraction of the entire set of learnable parameters. Moreover, TTT introduces additional layers for the unsupervised task which must be trained from scratch and could take a while for the training to stabilize, whereas ODR simply refines the already trained parameters.

The complete workflow of ODR is visualized in Figure \ref{fig:flowchart} along with the offline training and pose inference stages for comparison. The offline training updates the parameters of the entire network, including the shared feature encoder ($f$) and the prediction heads ($\Ehead, \Hhead, \Shead$), by minimizing the sum of losses from all heads, i.e., $\ell_E + \ell_H + \ell_S$. Then, at inference, only the outputs of $\Hhead$ and $\Ehead$ are used to predict the pose. Meanwhile, ODR does not involve $\Hhead$ and $\Ehead$, but instead minimizes the entropy loss $\ell_\text{ent}$ only through $\Shead$ on the target domain images. During ODR, the parameteres of the prediction heads are locked, and only the affine transformation parameters of the BN or GN layers in $f$ are updated.

\subsection{Entropy Minimization}

ODR minimizes Shannon entropy on $\Shead$. In classification tasks, entropy minimization encourages the predicted probability distribution to have distinct peaks, effectively enhancing the confidence in predicted classes. Mathematically, let $\bm{\theta}_f$ denote the set of all learnable parameters in the feature encoder, and $\bm{\theta}_S$ denote the same for the segmentation head. Let $\bm{\theta}_f^\text{norm} \subset \bm{\theta}_f$ denote the set of parameters associated with the normalization layers of the feature encoder. Then, the refinement procedure amounts to solving 
\begin{align}
	\min_{\bm{\theta}_f^\text{norm}} \frac{1}{n} \sum_{i=1}^n \ell_\text{ent}(\bm{x}_i; \bm{\theta}_f, \bm{\theta}_\text{S}),
\end{align}
where $\{\bm{x}_i\}$ are the unlabeled target domain images, and the entropy loss is given as
\begin{equation}
	\begin{aligned}
		\ell_\text{ent}(\bm{x}_i; \bm{\theta}_f, \bm{\theta}_\text{S}) = -\sum_p \sigma(\hat{\bm{y}}_{i,p}) \log \sigma(\hat{\bm{y}}_{i,p}),
	\end{aligned}
\end{equation}
where $\hat{\bm{y}}_i = \Shead(f(\bm{x}_i))$ is the output logit of the predicted segmentation map, $\hat{\bm{y}}_{i,p}$ is the $p$-th pixel of $\hat{\bm{y}}_i $, and $\sigma(\cdot)$ is a pixel-wise sigmoid function. 

\begin{figure*}[!t]
	\centering
	\includegraphics[width=\textwidth,trim=4 4 4 4,clip]{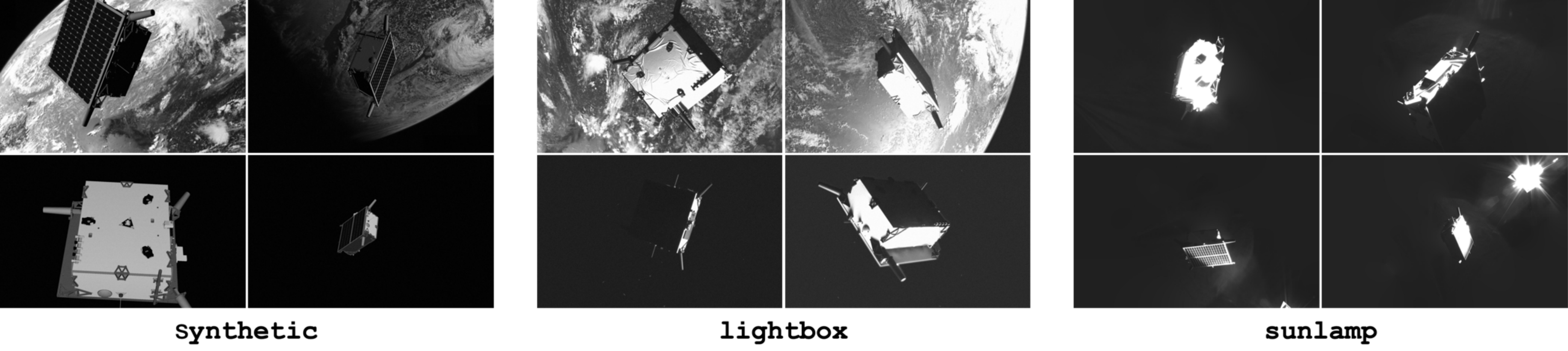}
	\caption{Example images from different domains of SPEED+. Figure from \citet{Park2021speedplus}.}
	\label{fig:speedplus overview}
\end{figure*}

\subsection{Normalization Layer Parameter Updates}

As explained, only the affine transformation parameters of the normalization layers of the feature encoder are updated through backpropagation during ODR. Formally, let $(\bm{z}_j^\text{in}, \bm{z}_j^\text{out})$ denote the input and output features of the $j$-th normalization layer in the feature encoder $f$. A normalization layer modulates and updates the feature according to  
\begin{align}
	\bm{z}_{j,c}^\text{out} = \text{Norm}(\bm{z}_{j,c}^\text{in}) = \gamma_{j,c} \bigg( \frac{\bm{z}_{j,c}^\text{in} - \mu}{\sigma} \bigg) + \beta_{j,c},	
\end{align}
where $\bm{z}_{j,c}^\text{in}$ denotes the input feature vector at $c$-th channel, $(\mu, \sigma^2)$ are the feature mean and variance applied across the batch of size $N$ for BN or across some $G$ channels for GN, and $(\gamma_{j,c}, \beta_{j,c})$ are the affine transformation parameters of the $j$-th normalization layer associated with $c$-th channel of the input feature. Then, the set of refinement parameters, $\bm{\theta}_f^\text{norm}$, consists of the affine parameters ($\gamma, \beta$) of all normalization layers in $f$, i.e., $\bm{\theta}_f^\text{norm} = \{\gamma_{j,c}, \beta_{j,c}\}$. 

In this work, if the BN layers are used to construct the encoder, then the feature mean and variance at each BN layer, which should approximate those of the features across the entire target domain, are updated online every $B$ images as running averages according to
\begin{equation} \label{eqn:update running averages}
	\begin{aligned}
		\mu_{(i+1)} &= m\mu_{(i)} + (1-m) \mu_B, \\
		\sigma^2_{(i+1)} &= m\sigma^2_{(i)} + (1-m)\sigma_B^2,
	\end{aligned}
\end{equation}
where $m \in [0, 1]$ is a momentum hyperparameter, $(\mu_{(i)}, \sigma_{(i)})$ are the current running averages of feature statistics, and $(\mu_B, \sigma_B^2)$ are the new observed feature statistics over the most recent $B$ images. The key difference against the conventional BN is that the samples are not processed in batches due to memory limits; they are instead processed sequentially, and the batch mean and variance are updated online \citep{Knuth1997ArtofComputerProgramming} as done for AdaBN \citep{Li2016AdaBN}.

\section{Experiments} \label{section:Experiments}

This section describes the experimental procedures and results for both offline robust training and online domain refinement.

\subsection{Dataset}
SPNv2 is trained and evaluated on SPEED+ \citep{Park2021speedplus, Park2021speedplusSDR} which comprises data from three distinct domains: $\synthetic$, $\lightbox$ and $\sunlamp$. The $\synthetic$ domain consists of 59,960 computer-generated images of the Tango spacecraft from the PRISMA mission \citep{PRISMA_chapter}. On the other hand, $\lightbox$ and $\sunlamp$ respectively contain 6,740 and 2,791 Hardware-In-the-Loop (HIL) images of the mockup model of the same spacecraft captured in the high-fidelity robotic simulation environment of the Testbed for Rendezvous and Optical Navigation (TRON) facility at the Space Rendezvous Laboratory (SLAB) of Stanford Univerisity. Specifically, the $\lightbox$ images are simulated with a set of calibrated lightboxes emulating the diffuse light in Earth's orbits, whereas the $\sunlamp$ images are illuminated with a metal halide arc lamp to mimic the direct sunlight commonly encountered in space. Figure \ref{fig:speedplus overview} provides visualization of a few samples from each domain. For more information on SPEED+, the readers are referred to \citet{Park2021speedplus}. In addition to SPEED+, the binary masks for the $\synthetic$ images are created in order to train SPNv2 for the segmentation task. SPNv2 is also evaluated on $\prisma$ which contains 25 labeled spaceborne images of Tango acquired during the rendezvous phase of the PRISMA mission \citep{PRISMA_chapter, Damico2014IJSSE}.

In this work, the $\synthetic$ domain's training and validation sets are used exclusively for offline robust training of SPNv2. Then, each $\lightbox$ and $\sunlamp$ \emph{unlabeled} images are used for ODR. This practice resembles the real-life mission constraint, where images from the target domain are used only when that domain becomes operationally available. The $\prisma$ images are not used for ODR and reserved only for evaluation of offline training since its quantity is severely limited for ODR.

\begin{table*}[!t]
	\caption{Bounding box and pose predictions of SPNv2 different head configurations. The performances from $\Hhead$ (H) and $\Ehead$ (E) on the SPEED+ \texttt{lightbox} and \texttt{sunlamp} domains are reported wherever applicable. {The performances from $\Hhead$ exclude the rejected outliers.} Bold numbers indicate best performances.}
	\label{tab:results_varying_heads}
	\centering
	\tabcolsep=0.12cm
	\begin{tabular}{@{}lccccccccccc@{}}
		\toprule
		\multirow{2}{*}{Heads} & \multirow{2}{*}{Source} & \multicolumn{5}{c}{\texttt{lightbox}} &  \multicolumn{5}{c}{\texttt{sunlamp}} \\
		\cmidrule(lr){3-7} \cmidrule(lr){8-12}
		&  & {\# Rej.} & IoU [-] & $E_\textrm{T}$ [m] & $E_\textrm{R}$ [$^\circ$] & $E_\textrm{pose}^*$ [-] & {\# Rej.} & IoU [-] & $E_\textrm{T}$ [m] & $E_\textrm{R}$ [$^\circ$] & $E_\textrm{pose}^*$ [-] \\
		\midrule
		E & E & {-} & {0.887} & {0.368} & {20.258} & {0.411} & {-} & {0.879} & {0.457} & {34.916} & {0.682} \\
		\midrule
		H & H & {2} & - & {0.235} & {10.074} & {0.214} & {3} & - & {0.292} & {15.957} & {0.327} \\
		\midrule
		\multirow{2}{*}{E + H} & E & - & {0.908} & {0.222} & {9.240} & {0.196} & - & {0.913} & {0.238} & {16.015} & {0.320} \\
		& H & {5} & - & {0.204} & {7.648} & {0.165} & {4} & - & {0.262} & {14.808} & {0.302} \\
		\midrule
		\multirow{2}{*}{E + H + S} & E & - & \textbf{0.918} & 0.175 & 8.004 & 0.169 & - & \textbf{0.919} & \textbf{0.225} & 12.433 & 0.254 \\
		& H & {3} & - & {\textbf{0.174}} & {\textbf{6.439}} & \textbf{0.158} & {4} & - & {0.238} & {\textbf{10.876}} & \textbf{0.245} \\
		\bottomrule
	\end{tabular}
\end{table*}

\begin{table*}[!t]
	\caption{Bounding box and pose predictions of full configuration SPNv2 and different data augmentation configurations. Starting with the baseline set of augmentations including Brightness \& Contrast, Blur and Noise, different augmentation techniques are progressively added to the training. {The performances from $\Hhead$ exclude the rejected outliers.} Bold numbers indicate best performances.}
	\label{tab:results_varying_augment}
	\centering
	\tabcolsep=0.1cm
	\begin{tabular}{@{}lccccccccccc@{}}
		\toprule
		\multirow{2}{*}{Config.} & \multirow{2}{*}{Source} & \multicolumn{5}{c}{\texttt{lightbox}} &  \multicolumn{5}{c}{\texttt{sunlamp}} \\
		\cmidrule(lr){3-7} \cmidrule(lr){8-12}
		&  & {\# Rej.} & IoU [-] & $E_\textrm{T}$ [m] & $E_\textrm{R}$ [$^\circ$] & $E_\textrm{pose}^*$ [-] & {\# Rej.} & IoU [-] & $E_\textrm{T}$ [m] & $E_\textrm{R}$ [$^\circ$] & $E_\textrm{pose}^*$ [-] \\
		\midrule
		\multirow{2}{*}{Baseline} & E & {-} & {0.863} & {0.481} & {23.638} & {0.485} & {-} & {0.744} & {1.366} & {61.177} & {1.262} \\
		& H & {6} & - & {0.427} & {21.048} & {0.434} & {28} & - & {1.599} & {60.535} & {1.298} \\ 
		\midrule
		\multirow{2}{*}{+ Random Erase} & E & {-} & {0.809} & {0.760} & {23.732} & {0.527} & {-} & {0.500} & {2.715} & {75.125} & {1.698} \\
		& H & {21} & - & {0.625} & {20.697} & {0.472} & {60} & - & {2.691} & {74.295} & {1.695} \\ 
		\midrule 
		\multirow{2}{*}{+ Sun Flare} & E & {-} & {0.905} & {0.259} & {10.357} & {0.221} & {-} & {0.857} & {0.618} & {28.191} & {0.586} \\
		& H & {5} & - & {0.242} & {8.577} & {0.186} & {9} & - & {0.670} & {27.033} & {0.571} \\
		\midrule
		\multirow{2}{*}{+ Style Aug.} & E & - & \textbf{0.918} & 0.175 & 8.004 & 0.169 & - & \textbf{0.919} & \textbf{0.225} & 12.433 & 0.254 \\
		& H & {3} & - & {\textbf{0.174}} & {\textbf{6.439}} & \textbf{0.158} & {4} & - & {0.238} & {\textbf{10.876}} & \textbf{0.245} \\
		\bottomrule
	\end{tabular}
\end{table*}

\subsection{Metrics}
For bounding boxes predicted by $\Ehead$, the standard IoU metric is used. For pose solutions regressed by $\Ehead$ and computed via Perspective-$n$-Point (P$n$P) \citep{Lepetit2008EPnP} from the keypoints predicted from $\Hhead$, rotation error ($E_\text{R}$) in degrees (Eq.~\ref{eqn:rotation error}), translation error ($E_\text{T}$) in meters (Eq.~\ref{eqn:translation error}), and SPEED score (Eq.~\ref{eqn:speed score}) are reported. Note that when evaluated on SPEED+ $\lightbox$ and $\sunlamp$ samples, the HIL pose error ($E_\text{pose}^*$) is used to zero out the errors smaller than the thresholds based on the TRON calibration \citep{park2023spec2021}. For an individual sample, define the HIL translation error ($E_\text{t}^*$) and HIL rotation error ($E_\text{R}^*$) as
\begin{subequations}
\begin{align}
    E_\text{t}^*(\tilde{\bm{t}}, \bm{t}) &= \begin{cases} 0 & \textrm{if } E_\text{t}(\tilde{\bm{t}}, \bm{t}) / \|\bm{t}\| < 2.173 \textrm{ mm/m} \\ E_\text{t}(\tilde{\bm{t}}, \bm{t}) & \textrm{otherwise} \end{cases}, \\ E_\text{R}^*(\tilde{\bm{R}}, \bm{R}) &= \begin{cases} 0 & \textrm{if } E_\text{R}(\tilde{\bm{R}}, \bm{R}) < 0.169^\circ \\ E_\text{R}(\tilde{\bm{R}}, \bm{R}) & \textrm{otherwise} \end{cases},
\end{align}
\end{subequations}
where $\|\bm{t}\|$ is the ground-truth distance to the target. Then, the HIL pose error is
\begin{align}
	E_\text{pose}^* = E_\text{R}^*(\tilde{\bm{R}}, \bm{R}) + E_\text{t}^*(\tilde{\bm{t}}, \bm{t}) / \|\bm{t}\|.
\end{align}

It is important to note that $\Hhead$ and P$n$P would often return outlier predictions on some HIL images due to significant domain gap and noisy heatmap predictions. A few of these would particularly blow up the mean translation error. In order to prevent this, the experiment results report the number of outliers predicted from $\Hhead$ which are excluded from computing the mean $E_\text{t}^*$. Specifically, an outlier is defined as a translation vector which lies outside a viewing frustum of the camera defined by the minimum distance of 0 m, maximum distance of 50 m and its field of view. While the SPEED+ position distribution is within a 10 m separation, 50 m is chosen as it approximately matches the distance at which the target becomes unresolved for the camera used in SPEED+. This criterion is only applied to the translation vectors from $\Hhead$, not $\Ehead$.

\subsection{Implementation Details} \label{sec:experiments_implementations}

In all offline experiments on the $\synthetic$ domain, SPNv2 is trained with the AdamW \citep{Loshchilov2017AdamW} optimizer with the initial learning rate of $1 \times 10^{-3}$ which decays by the factor of 0.1 after 15 and 18 epochs. The training lasts 20 epochs (i.e., trains on the entire training set 20 times) for all but a configuration with only $\Hhead$, which is trained for only 10 epochs since heatmap prediction is easier to learn. The input image is resized to 768 $\times$ 512, which minimizes the loss of information due to excessive downscaling and has an approximately matching aspect ratio compared to the original 1920 $\times$ 1200 images. Note that unlike existing methods for spacecraft pose estimation \citep{Park2019AAS, Sharma2019AAS, Chen2019SatellitePE}, no separate object detection network is used to detect and crop out the region-of-interest around the spacecraft in advance before feeding it to the pose estimation network. The reason is that, unlike SPEED whose farthest image is nearly 40 m away, SPEED+ data are limited to maximum 10 m separation, so the shape of the farthest spacecraft can still be recognized at high enough image resolution.

Unless noted otherwise, the EfficientNet scaling parameter is set to $\phi = 3$, which amounts to roughly 12M parameters for the EfficientNet-B3 backbone and BiFPN layers, and BN layers are used to construct the feature encoder. All training is run on 4 NVIDIA V100 GPUs with per-GPU batch size 4 except for $\phi = 6$ which is trained with per-GPU batch size 1. When trained on multiple GPUs with BN layers, its statistics are synchronized across devices.

For ODR, the test images are provided one at a time with no data augmentation. The running mean and variance of each BN layers are updated after every $B$ images with $m$ = 0.9, and the experiment terminates after SPNv2 has seen $N$ random samples from the test domain. ODR experiments are conducted on a single GPU.

\subsection{Results: Offline Robust Training}

\begin{table*}[!t]
	\caption{Comparison of SPNv2 models with different scaling parameter $\phi$ and normalization layers in the feature encoder. The number of parameters counts those of the feature encoder. Bold numbers indicate best performances. BN: batch normalization. GN: group normalization.}
	\label{tab:backbone_normalization_comp}
	\centering
	\tabcolsep=0.1cm
	\begin{tabular}{@{}ccccccccccc@{}}
		\toprule
		\multirowcell{2}{$\phi$ \\ (Num. of Param.)} &  \multirowcell{2}{Norm. \\ Layer} & \multicolumn{3}{c}{\texttt{synthetic}} &  \multicolumn{3}{c}{\texttt{lightbox}} & \multicolumn{3}{c}{\texttt{sunlamp}} \\
		\cmidrule(lr){3-5} \cmidrule(lr){6-8} \cmidrule(lr){9-11}
		& & $E_\textrm{T}$ [m] & $E_\textrm{R}$ [$^\circ$] & $E_\textrm{pose}$ [-] & $E_\textrm{T}$ [m] & $E_\textrm{R}$ [$^\circ$] & $E_\textrm{pose}^*$ [-] & $E_\textrm{T}$ [m] & $E_\textrm{R}$ [$^\circ$] & $E_\textrm{pose}^*$ [-] \\
		\midrule
		$\phi$ = 0 (3.8M) & BN & 0.075 & 1.149 & 0.033 & 0.355 & 12.108 & 0.268 & 0.349 & 18.734 & 0.385 \\
		$\phi$ = 3 (12.0M) & BN & 0.054 & 0.987 & 0.027 & \bfseries 0.175 & \bfseries 6.479 & \bfseries 0.142 & \bfseries 0.225 & 11.065 &  0.230 \\
		$\phi$ = 3 (12.0M) & GN & 0.056 & 1.224 & 0.031 & 0.247 & 12.919 & 0.267 & 0.342 & 28.041 & 0.545  \\
		$\phi$ = 6 (52.5M) & GN & \bfseries 0.031 & \bfseries 0.885 & \bfseries 0.021 & 0.216 & 7.984 & 0.173 & 0.230 & \bfseries 10.369 & \bfseries 0.219  \\
		\bottomrule
	\end{tabular}
\end{table*}

First, SPNv2 is trained offline with a full set of data augmentations listed in Table \ref{tab:data augmentations} but with different configurations of prediction heads. Table \ref{tab:results_varying_heads} shows that for a full set of data augmentations applied, having more prediction heads in the architecture improves the CNN robustness on the SPEED+ HIL domains. The performance culminates with the full configuration visualized in Figure \ref{fig:architecture}, which hints that jointly training a shared feature extractor for different tasks prevents it from learning task-specific features. On the other hand, Table \ref{tab:results_varying_augment} fixes the SPNv2 architeture to full configuration and instead studies the effects of three data augmentation techniques: Random Erase \citep{Zhong2020RandomErasing}, Sun Flare, and Style Augmentation \citep{Jackson2019ICCV_StyleAug}. When random erase is applied alone, the pose error actually increases {for the $\sunlamp$ domain}, almost doubling for the $\sunlamp$ domain, suggesting the occlusion introduced by the random erase makes the learning much harder. However, as the other two augmentations are progressively added to the training, the pose error also decreases. Interestingly, adding sun flare augmentation significantly improves the pose errors on {both domains} the $\lightbox$ domain compared to the baseline, whereas adding style augmentation does so on the $\sunlamp$ domain.

An interesting observation is that, even at the optimal configuration with low pose errors on HIL domains, there are still outliers that would otherwise make $E_\text{t}^*$ unacceptably large. Excluding these outliers from $\Hhead$, the accuracy of the translation vectors regressed from $\Ehead$ is generally on par with that of the inliers from $\Hhead$. This suggests that, in the absence of a natural outlier rejection scheme from, e.g., a navigation filter, the translation vectors from $\Ehead$ are more robust than those from $\Hhead$. On the other hand, the orientation errors are consistently lower from $\Hhead$ regardless of the presence of outliers. Based on this observation, the pose predictions of the full configuration SPNv2 onward report translation error from $\Ehead$ and orientation error from $\Hhead$ from \emph{all} samples unless otherwise noted.

Finally, the EfficientNet scaling parameter $\phi$, which controls the number of parameters associated with the feature encoder, and the normalization layers (BN or GN) are varied to gauge their effect on the CNN's generalization capability across domain gaps. The SPNv2 is trained with same training parameters for different size and normalization layer configurations. The results shown in Table \ref{tab:backbone_normalization_comp} indicate that, while there is no noticeable difference in performance on the $\synthetic$ validation set, the performances on the SPEED+ HIL domains differ considerably depending on the network size and the type of normalization layers used to build the feature encoder. Specifically, compared to the generalizability accomplished by the default architecture ($\phi$ = 3, BN), reducing the network size to $\phi = 0$ or replacing the BN layers with GN layers significantly degrades the network's performance on the HIL test sets. In order to match the performance of the default architecture with a completely batch-agnostic one, the network size must be increased to $\phi = 6$, which more than quadruples the size of the feature encoder. The trend observed in Table \ref{tab:backbone_normalization_comp} is a direct opposite of the desired characteristics of CNN models for spaceborne computer vision applications elaborated in Section \ref{section:ODR}, which aim for a smaller, batch-agnostic neural network architecture suitable for the on-board satellite avionics. The search for such an architecture is left as a future work.

\subsection{Results: Online Domain Refinement}

\begin{figure*}[!p]
	\centering
	\begin{subfigure}[b]{0.48\textwidth}
		\centering
		\includegraphics[width=\textwidth]{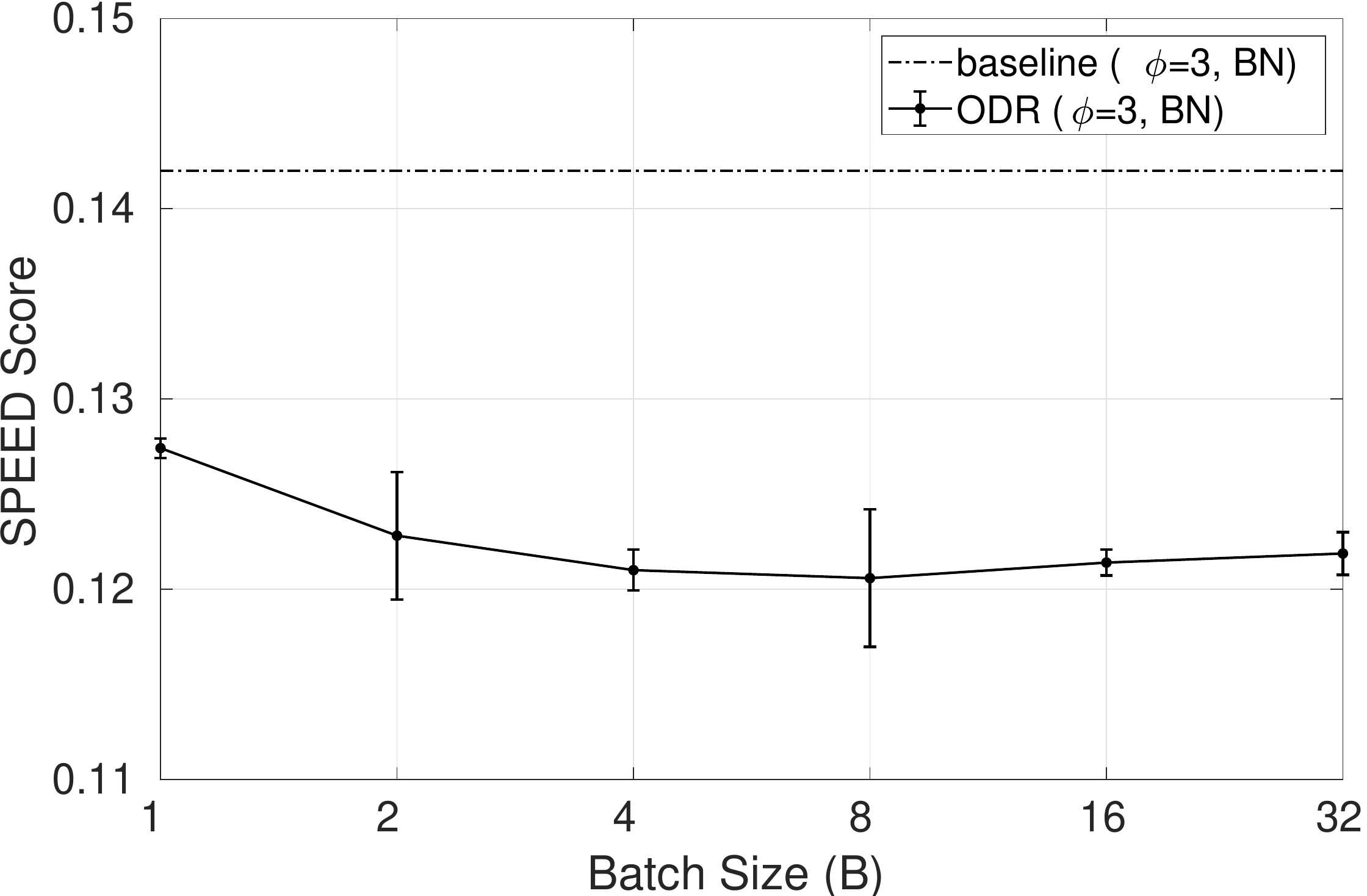}
		\caption{ODR on $\lightbox$}
		\label{fig:ttdr_vary_batch_lightbox}
	\end{subfigure}
	\begin{subfigure}[b]{0.48\textwidth}
		\centering
		\includegraphics[width=\textwidth]{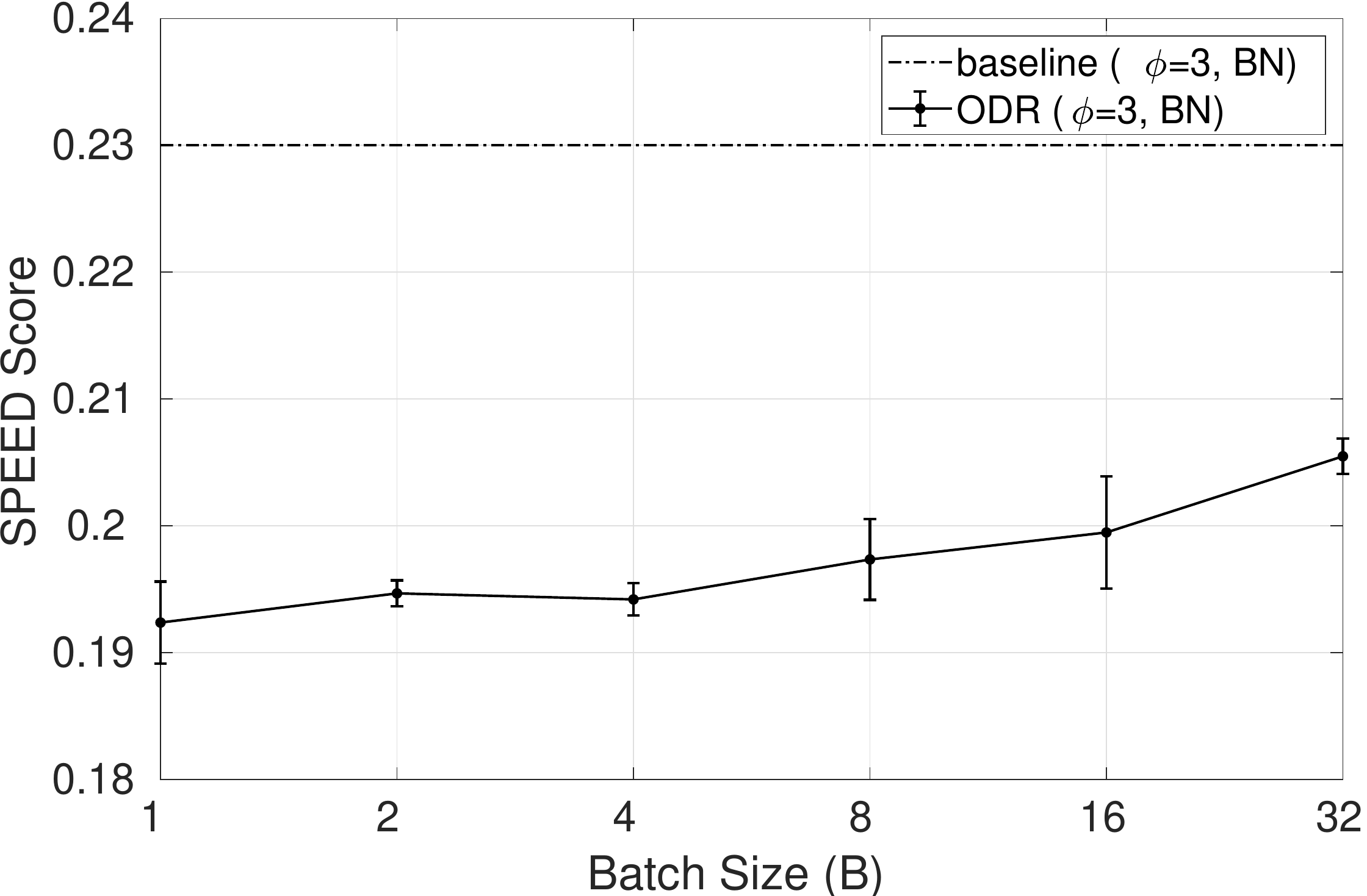}
		\caption{ODR on $\sunlamp$}
		\label{fig:ttdr_vary_batch_sunlamp}
	\end{subfigure}
	\caption{ODR on SPEED+ test domains with varying batch size ($B$). Total number of samples seen is fixed to $N$ = 1024. Error bars represent standard deviations over 5 runs of ODR with different random seeds.}
	\label{fig:ttdr_vary_batch}
\end{figure*}

\begin{figure*}[!p]
	\centering
	\begin{subfigure}[b]{0.48\textwidth}
		\centering
		\includegraphics[width=\textwidth]{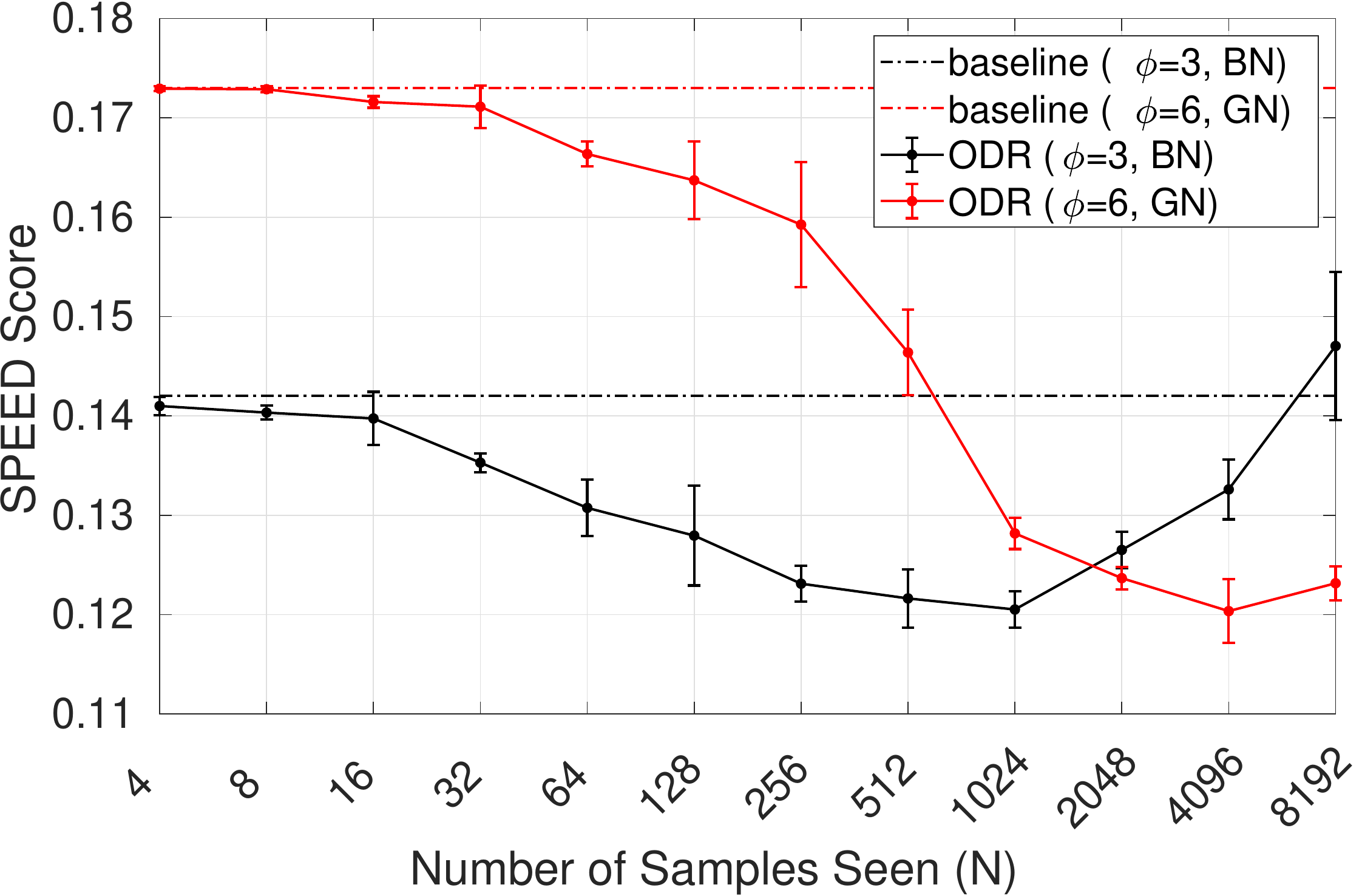}
		\caption{ODR on $\lightbox$}
		\label{fig:ttdr_vary_sample_lightbox}
	\end{subfigure}
	\begin{subfigure}[b]{0.48\textwidth}
		\centering
		\includegraphics[width=\textwidth]{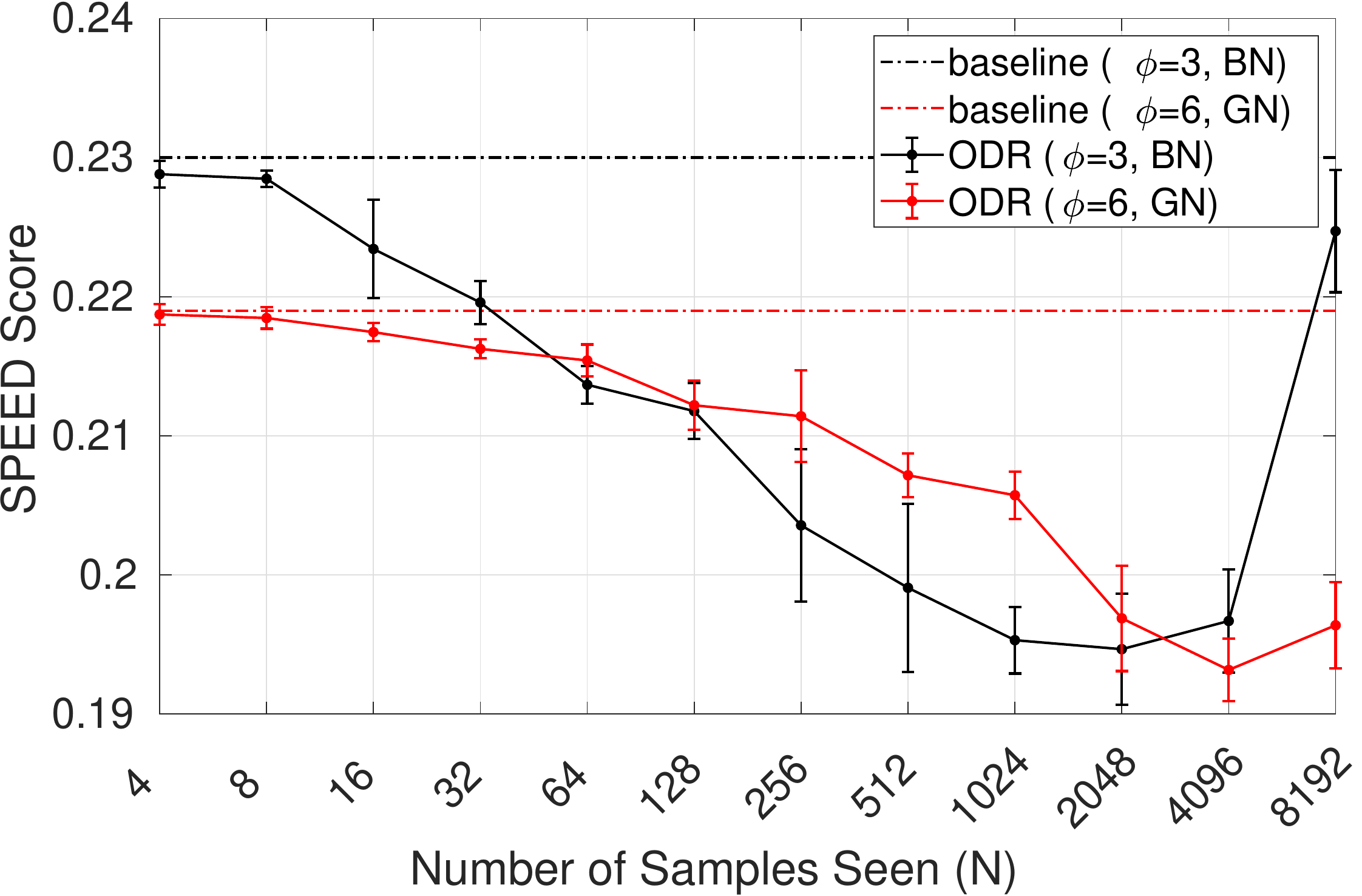}
		\caption{ODR on $\sunlamp$}
		\label{fig:ttdr_vary_sample_sunlamp}
	\end{subfigure}
	\caption{ODR on SPEED+ test domains with varying number of observed samples ($N$). The batch size is fixed to $B = 4$. Error bars represent standard deviations over 5 runs of ODR with different random seeds.}
	\label{fig:ttdr_vary_sample}
\end{figure*}

\begin{figure*}[!p]
	\centering
	\begin{subfigure}[b]{0.48\textwidth}
		\centering
		\includegraphics[width=\textwidth]{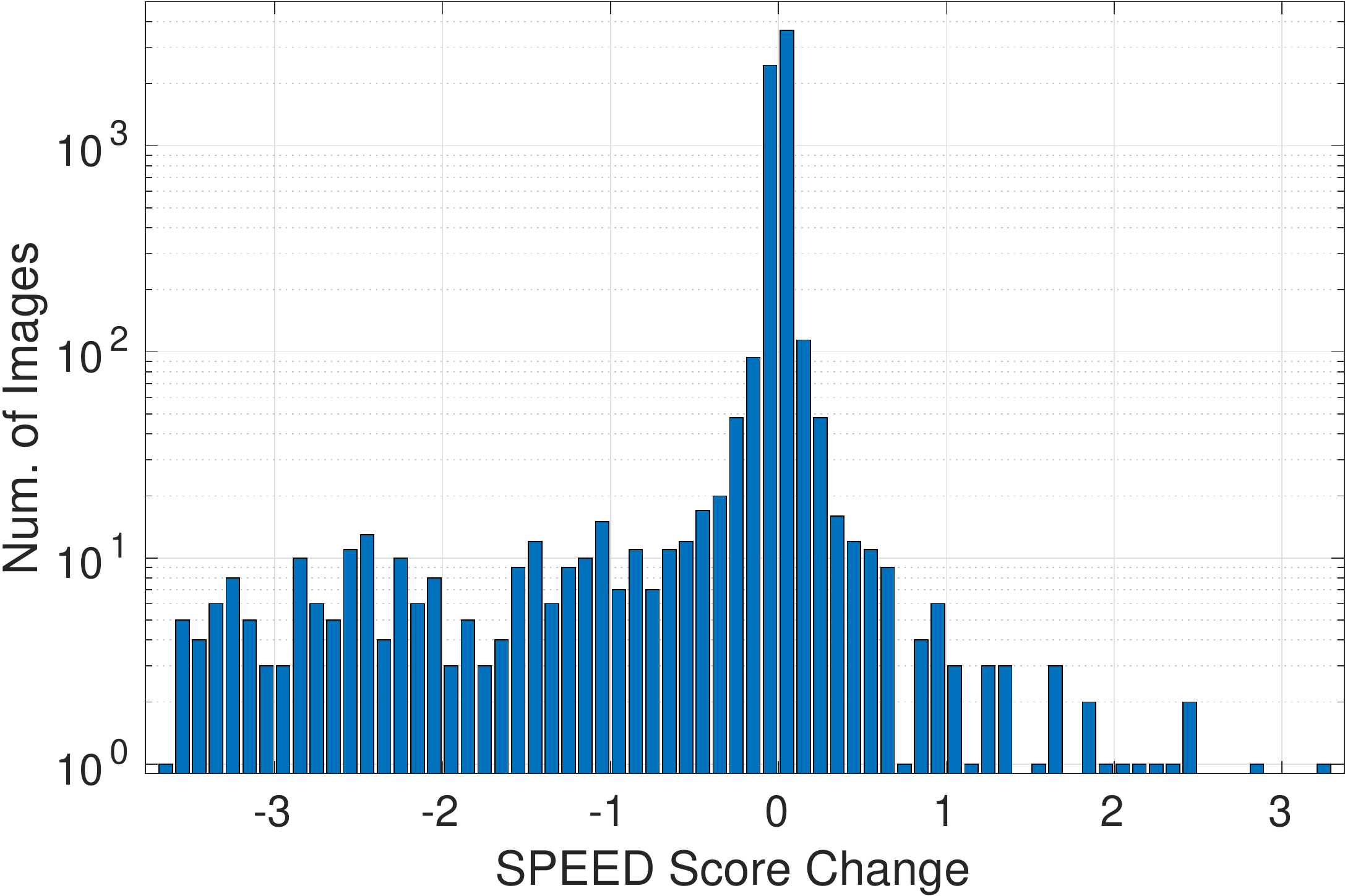}
		\caption{SPEED score change on $\lightbox$}
		\label{fig:score_change_lightbox}
	\end{subfigure}
	\begin{subfigure}[b]{0.48\textwidth}
		\centering
		\includegraphics[width=\textwidth]{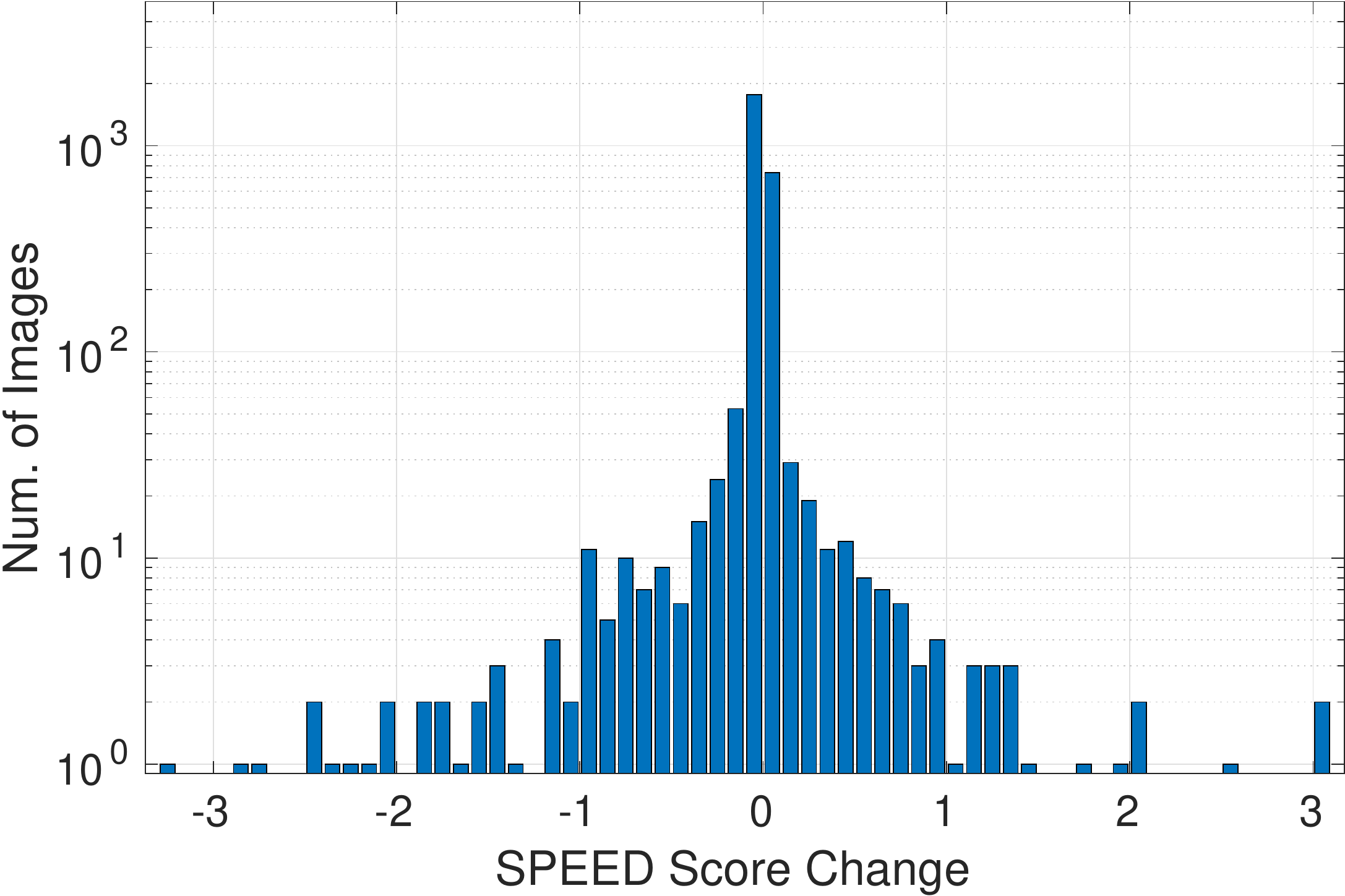}
		\caption{SPEED score change on $\sunlamp$}
		\label{fig:score_change_sunlamp}
	\end{subfigure}
	\caption{Distribution of change in SPEED score ($E^*_\text{pose}$) of the SPNv2 ($\phi = 6$, GN) after ODR with $N$ = 4096 on SPEED+ HIL domains. Number of images in $y$-axis is shown in log scale.}
	\label{fig:score_change}
\end{figure*}

\begin{figure*}[!p]
	\centering
	\includegraphics[width=\textwidth]{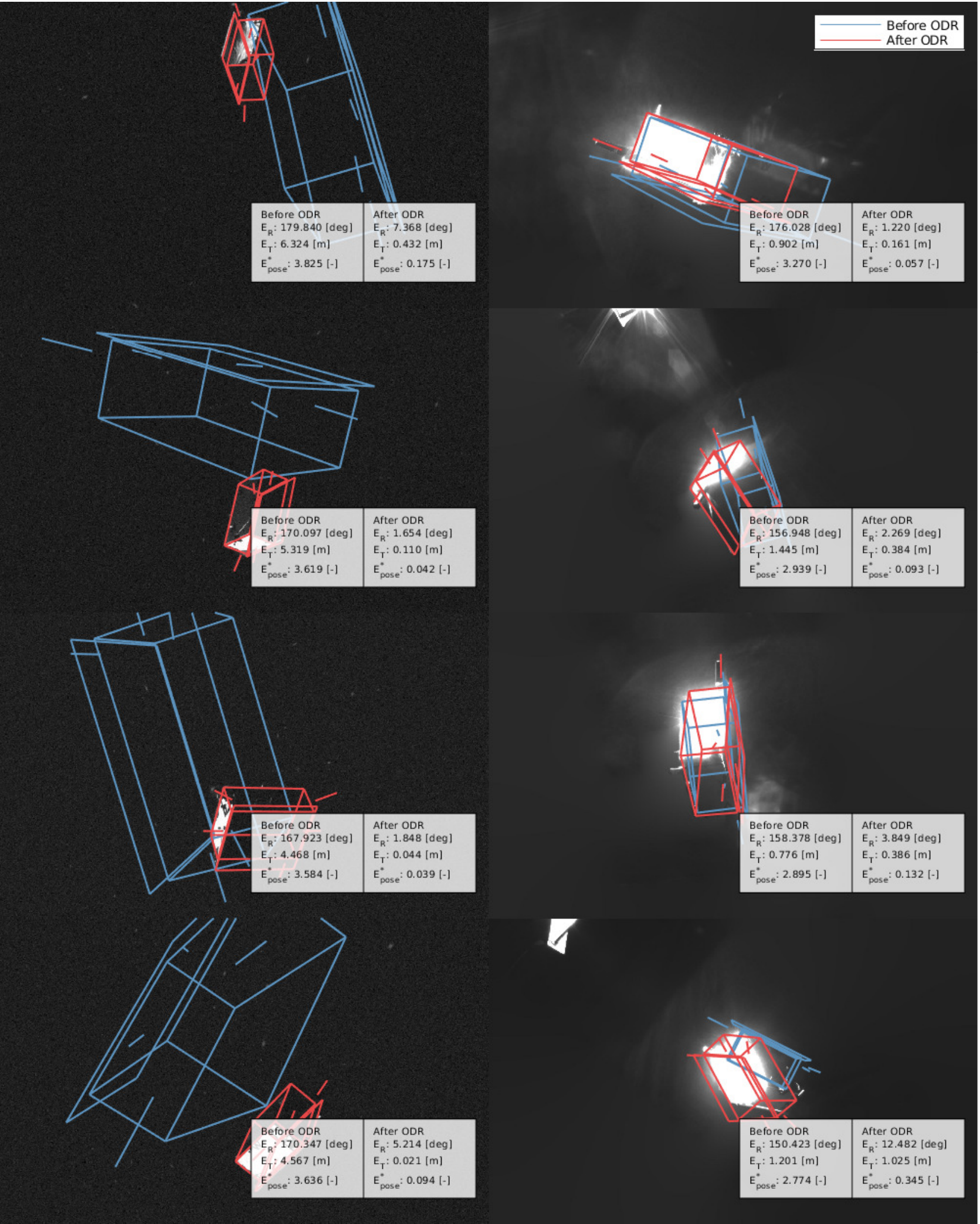}
	\caption{Visualization of examples on which the pose predictions of SPNv2 significantly improve after ODR on $\lightbox$ (\emph{left}) and $\sunlamp$ (\emph{right}) domains.}
	\label{fig:improvement}
\end{figure*}

The ODR experiments in this section build upon the full configuration SPNv2 trained offline with the full set of augmentations, which is considered as the baseline. Based on the results in Table \ref{tab:backbone_normalization_comp}, two SPNv2 architectures are considered: small but batch-dependent ($\phi$ = 3, BN) and large but batch-agnostic ($\phi$ = 6, GN). Recall that ODR has two hyperparameters:
\begin{itemize}
	\item $B$, the image batch size which controls the frequency of updating the running average of the BN layer statistics according to Eq.~\ref{eqn:update running averages}, and
	\item $N$, the total number of target domain images the neural network observes during ODR.
\end{itemize}

\begin{table*}[!t]
	\caption{Comparison of final poses predicted by SPNv2 and KRN \citep{Park2019AAS} from the SPEED+ baseline studies \citep{Park2021speedplus}. KRN's oracle performance denotes mean error from 5-fold cross validation on the respective SPEED+ HIL domains. ODR is not run on $\prisma$ due to limited available samples. Bold numbers indicate best performances excluding oracle.}
	\label{tab:final_results_offline}
	\centering
	\tabcolsep=0.1cm
	\begin{tabular}{@{}lccccccccc@{}}
		\toprule
		\multirow{2}{*}{CNN Architecture} & \multicolumn{3}{c}{\texttt{lightbox}} &  \multicolumn{3}{c}{\texttt{sunlamp}} & \multicolumn{3}{c}{\texttt{prisma25}} \\
		\cmidrule(lr){2-4} \cmidrule(lr){5-7} \cmidrule(lr){8-10}
		& $E_\textrm{T}$ [m] & $E_\textrm{R}$ [$^\circ$] & $E_\textrm{pose}^*$ [-] & $E_\textrm{T}$ [m] & $E_\textrm{R}$ [$^\circ$] & $E_\textrm{pose}^*$ [-] & $E_\textrm{T}$ [m] & $E_\textrm{R}$ [$^\circ$] & $E_\textrm{pose}$ [-] \\
		\midrule
		KRN \citep{Park2019AAS} & 2.25 & 44.53 & 1.12 & 14.64 & 80.95 & 3.73 & 2.64 & 86.04 & 1.76 \\
		+ Style augmentation& 1.06 & 36.14 & 0.81 & 1.32 & 62.85 & 1.32 & 4.06 & 20.57 & 0.71 \\
		Oracle & 0.24 & 6.15 & 0.15 & 0.19 & 5.33 & 0.13 & - & - & - \\
		\midrule
		\underline{SPNv2 (Offline)} &&&&&&&&& \\
		- $\phi$ = 3, BN & 0.175 & 6.479 & 0.142 & 0.225 & 11.065 & 0.230 & 1.572 & \bfseries 5.202 & \bfseries 0.216 \\
		- $\phi$ = 6, GN & 0.216 & 7.984 & 0.173 & 0.230 & 10.369 & 0.219 & \bfseries 1.570 & 9.014 & 0.283 \\
		\midrule
		\underline{SPNv2 (ODR)} &&&&&&&&& \\
		- $\phi$ = 3, BN ($B = 4, N = 1024$) & \bfseries 0.142 &  5.624 & \bfseries 0.122 & 0.182 & \bfseries 9.601 & 0.198 & - & - & - \\
		- $\phi$ = 6, GN ($N = 4096$)  & 0.150 & \bfseries 5.577 & 0.122 & \bfseries 0.161 & 9.788 & \bfseries 0.197 & - & - & - \\
		\bottomrule
	\end{tabular}
\end{table*}

The hyperparameter $B$ is irrelevant if the network only consists of the GN layers. In this section, the effect of the ODR batch size $B$ is first tested on the batch-dependent architecture of SPNv2. The results are shown in Figure \ref{fig:ttdr_vary_batch}, where $N$ = 1024 is fixed. It shows that there is a benefit of increasing the batch size to $B = 4$ or $B = 8$ for $\lightbox$, whereas there is no benefit of updating the BN statistics after more than $B = 1$ image for $\sunlamp$. From this observation, the batch size is fixed to $B = 4$, i.e., the BN layer means and variances are updated after every 4 images, as it seems to strike the balance between both HIL domains. 

Then, the total number of observed samples ($N$) is varied for both batch-dependent and batch-agnostic architectures of SPNv2, as shown in Figure \ref{fig:ttdr_vary_sample}. As expected, as the unsupervised entropy minimization continues on, the SPEED score improves for both HIL domains. However, for the batch-dependent architecture, the improvement not only halts after $N$ = 1024 images for $\lightbox$ and $N$ = 2048 samples for $\sunlamp$, but it also becomes worse than the baseline performance after nearly 8000 images for the $\lightbox$ domain. On the other hand, for the batch-agnostic architecture, such failure is not observed for both HIL domains. In fact, the level of performance improvement is much larger for the batch-agnostic architecture on $\lightbox$. The most likely reason is that the ODR for the batch-agnostic architecture need not estimate the batch-wise feature normalization statistics which approximate those of the entire image domain. Therefore, the process is subject to significantly less uncertainty compared to that of the batch-dependent architecture.

The final scores of SPNv2 on SPEED+ and $\prisma$ are compared with KRN \citep{Park2019AAS} from the SPEED+ baseline studies \citep{Park2021speedplus} in Table \ref{tab:final_results_offline}. Based on Figure \ref{fig:ttdr_vary_sample}, the ODR is performed with $B=4, N=1024$ for the batch-dependent architecture and $N=4096$ for the batch-agnostic one. The pose predictions for $\prisma$ are from $\Hhead$ only since its images are captured with a camera different from that of SPEED+. It shows that not only SPNv2 with just offline training is more accurate than KRN trained with and without style augmentation, but also its performance on $\lightbox$ is even better than the oracle performance of KRN, i.e., when trained directly on the $\lightbox$ images and labels. In other words, SPNv2 can generalize better to \emph{unseen} $\lightbox$ images than KRN can after training on those exact images. It emphasizes a significant improvement of SPNv2 over KRN especially in its pose estimation architecture. SPNv2 also shows better performance on 25 spaceborne images of $\prisma$ compared to KRN for both variant of architectures. Finally, ODR for both variants of SPNv2 architecture reports around 15cm / 17cm translation errors and 5.6$^\circ$ /  $9.7^\circ$ orientation errors for $\lightbox$ and $\sunlamp$, respectively. While there is no noticeable difference in the reported values, the ODR on the batch-agnostic SPNv2 closes a much bigger performance gap on $\lightbox$ with no observable trend of failure of prediction after many iterations.

Finally, for the batch-agnostic SPNv2, Figure \ref{fig:score_change} shows the distribution of the changes in SPEED score after applying ODR, where a negative number suggests the score improved, and vice versa. For $\lightbox$, a vast majority of images enjoys improvement in SPEED score after ODR, with the largest score drop surpassing 3. For $\sunlamp$, the improvement is difficult to notice from the histogram, as significant number of images actually become more difficult to predict after ODR. Based on this result, the pose predictions of 4 samples that experience the largest improvement in SPEED score are visualized in Figure \ref{fig:improvement}.

\section{Discussion} \label{sec:limitations}

\begin{table}[!h]
	\caption{Number of tuned vs.~entire learnable parameters in the feature encoder of SPNv2.}
	\label{tab:number of parameters}
	\centering
	\tabcolsep=0.1cm
	\begin{tabular}{@{}lccc@{}}
		\toprule
		Architecture & Total & Tuned & Percentage (\%) \\
		\midrule
		$\phi$ = 3, BN & 12.0M & 106K & 0.88 \\
		$\phi$ = 6, GN & 52.5M & 287K & 0.55 \\
		\bottomrule
	\end{tabular}
\end{table}

The proposed ODR is source-free, online and computationally efficient. Specifically, as shown in Table \ref{tab:number of parameters}, the number of parameters tuned in the process of ODR is less than 1\% of all learnable parameters in either variant of SPNv2. However, there are some limitations that need to be overcome in future work for the proposed method to be fully mission-compliant.

First, the study of offline training of SPNv2 reveals that the good generalizability across domain gaps is best achieved by a larger network with batch-dependent normalization layers such as BN. Note that the smaller variant of SPNv2 ($\phi = 3$, BN) already takes on average 0.63 seconds for inference on an Intel\textregistered~Core\texttrademark~i9-9900K CPU at 3.60GHz and 1.05 seconds for an iteration of ODR, which consists of inference, gradient backpropagation and online update step. Therefore, a bigger variant becomes prohibitively large for frequent inference and ODR updates on-board the satellite hardware. This is a problem, since the study on ODR clearly indicates that the batch-agnostic architecture performs better in the absence of the need to estimate the domain-wise feature normalization statistics. Therefore, a future work must aim to search for a smaller and more efficient CNN architecture that is batch-agnostic and can generalize its performance across domain gaps at least as well as SPNv2 can.

The second limitation is that, while ODR improves the average performance on the HIL domain images as a whole, its predictions still worsen significantly on many images as shown in Figure \ref{fig:score_change}. An ideal refinement procedure would instead only introduce a non-negative change in the network's prediction accuracy on all images. This would require modifying the objective of the unsupervised learning or regularizing the loss function in such a way that prevents significant degradation of performance on individual images.

\section{Conclusions}

This paper presents Spacecraft Pose Network v2 (SPNv2), a multi-scale, multi-task convolutional neural network for pose estimation of noncooperative spacecraft across domain gaps. The SPNv2 architecture and its training mechanism are designed to comply with the realistic and operational constraints of machine learning in spaceborne applications. Specifically, SPNv2 is first trained offline on $\synthetic$ images of the SPEED+ dataset with extensive data augmentation and domain randomization. Then, at deployment, it refines the affine parameters of its feature extractor on unlabeled SPEED+ $\lightbox$ and $\sunlamp$ domains by self-supervised entropy minimization. It is shown that the combination of offline training and Online Domain Refinement (ODR) renders SPNv2 capable of bridging the domain gap, as evidenced by $0.15$ m and $5.58^\circ$ errors on $\lightbox$ and $0.16$ m and $9.79^\circ$ errors on $\sunlamp$ for a larger, batch-agnostic variant of SPNv2. Moreover, the proposed ODR is source-free, online, and computationally efficient, as it only refines less than 1\% of the learnable parameters in the feature extractor of SPNv2. The paper discusses a number of limitations of SPNv2 and ODR that must be addressed to build a fully mission-compliant model, which will be fully batch-agnostic, requires less computation, and does not deteriorate the prediction on the target domain samples throughout ODR. Overall, the methodology presented in this paper can easily extend to any spaceborne images of a known, noncooperative target collected in future missions. In the future, SPNv2 and ODR will be integrated into a navigation filter (e.g., unscented Kalman filter) to assess its capacity as an image processing unit of the filter in a number of close-range rendezvous scenarios. The uncertainty of the SPNv2's predictions and its impact on the filter performance will also be thoroughly evaluated.

\section{Acknowledgement}
This work is partially supported by Taqnia International through contract \#1232617-1-GWNDV. We would like to thank OHB Sweden for the 3D model of the Tango spacecraft used to create the images used in this article and for the flight images in the \texttt{prisma25} dataset. We acknowledge the Advanced Concepts Team (ACT) of the European Space Agency (ESA) for their contribution to putting together the SPEED+ dataset. We would also like to thank the Stanford Research Computing Center for providing computational resources that contributed to these research results.

\bibliographystyle{jasr-model5-names}
\biboptions{authoryear}
\bibliography{reference.bib}  

\begin{thebibliography}{54}
\expandafter\ifx\csname natexlab\endcsname\relax\def\natexlab#1{#1}\fi
\ifx\xfnm\relax \def\xfnm[#1]{\unskip,\space#1}\fi

\bibitem[{Ben-David et~al.(2010)Ben-David, Blitzer, Crammer, Kulesza, Pereira
  \& Vaughan}]{BenDavid2010LearningFromDiffDomains}
\bibinfo{author}{Ben-David, S.}, \bibinfo{author}{Blitzer, J.},
  \bibinfo{author}{Crammer, K.} et~al. (\bibinfo{year}{2010}).
\newblock \bibinfo{title}{A theory of learning from different domains}.
\newblock {\it \bibinfo{journal}{Mach.~Learn.}\/},  {\it
  \bibinfo{volume}{79}\/}, \bibinfo{pages}{151–175}.
  \DOIprefix\doi{10.1007/s10994-009-5152-4}.

\bibitem[{Black et~al.(2021)Black, Shankar, Fonseka, Deutsch, Dhir \&
  Akella}]{Black2021PoseEstimationCygnus}
\bibinfo{author}{Black, K.}, \bibinfo{author}{Shankar, S.},
  \bibinfo{author}{Fonseka, D.} et~al. (\bibinfo{year}{2021}).
\newblock \bibinfo{title}{Real-time, flight-ready, non-cooperative spacecraft
  pose estimation using monocular imagery}.
\newblock In {\it \bibinfo{booktitle}{31st AAS/AIAA Space Flight Mechanics
  Meeting}\/}.

\bibitem[{Bukschat \& Vetter(2020)}]{Bukschat2020EfficientPose}
\bibinfo{author}{Bukschat, Y.},  \& \bibinfo{author}{Vetter, M.}
  (\bibinfo{year}{2020}).
\newblock \bibinfo{title}{Efficientpose: An efficient, accurate and scalable
  end-to-end 6d multi object pose estimation approach}.
\newblock {\it \bibinfo{journal}{CoRR}\/},  {\it
  \bibinfo{volume}{abs/2011.04307}\/}.
  \href{http://arxiv.org/abs/2011.04307}{\tt arXiv:2011.04307}.

\bibitem[{Buslaev et~al.(2020)Buslaev, Iglovikov, Khvedchenya, Parinov,
  Druzhinin \& Kalinin}]{Buslaev2020Albumentations}
\bibinfo{author}{Buslaev, A.}, \bibinfo{author}{Iglovikov, V.~I.},
  \bibinfo{author}{Khvedchenya, E.} et~al. (\bibinfo{year}{2020}).
\newblock \bibinfo{title}{Albumentations: Fast and flexible image
  augmentations}.
\newblock {\it \bibinfo{journal}{Information}\/},  {\it
  \bibinfo{volume}{11}\/}\bibinfo{issue}{(2)}.
  \DOIprefix\doi{10.3390/info11020125}.

\bibitem[{Caruana(1997)}]{Caruana1997MultitaskLearning}
\bibinfo{author}{Caruana, R.} (\bibinfo{year}{1997}).
\newblock \bibinfo{title}{Multitask learning}.
\newblock {\it \bibinfo{journal}{Mach.~Learn.}\/},  {\it
  \bibinfo{volume}{28}\/}, \bibinfo{pages}{41–75}.
  \DOIprefix\doi{10.1023/A:1007379606734}.

\bibitem[{Chen et~al.(2019)Chen, Cao, Bustos \& Chin}]{Chen2019SatellitePE}
\bibinfo{author}{Chen, B.}, \bibinfo{author}{Cao, J.}, \bibinfo{author}{Bustos,
  {\'A}.~P.} et~al. (\bibinfo{year}{2019}).
\newblock \bibinfo{title}{Satellite pose estimation with deep landmark
  regression and nonlinear pose refinement}.
\newblock {\it \bibinfo{journal}{2019 IEEE/CVF International Conference on
  Computer Vision Workshop (ICCVW)}\/},  (pp. \bibinfo{pages}{2816--2824}).
  \DOIprefix\doi{10.1109/ICCVW.2019.00343}.

\bibitem[{Chen et~al.(2020)Chen, Kornblith, Norouzi \&
  Hinton}]{Chen2020ICMLContrastiveLearning}
\bibinfo{author}{Chen, T.}, \bibinfo{author}{Kornblith, S.},
  \bibinfo{author}{Norouzi, M.} et~al. (\bibinfo{year}{2020}).
\newblock \bibinfo{title}{A simple framework for contrastive learning of visual
  representations}.
\newblock In \bibinfo{editor}{H.~Daum\'{e}~III}, \& \bibinfo{editor}{A.~Singh}
  (Eds.), {\it \bibinfo{booktitle}{Proceedings of the 37th International
  Conference on Machine Learning}\/} (pp. \bibinfo{pages}{1597--1607}).
\newblock \bibinfo{publisher}{PMLR} volume \bibinfo{volume}{119} of {\it
  \bibinfo{series}{Proceedings of Machine Learning Research}\/}.
\newblock \URLprefix \url{https://proceedings.mlr.press/v119/chen20j.html}.

\bibitem[{D'Amico et~al.(2014)D'Amico, Benn \& J{\o}rgensen}]{Damico2014IJSSE}
\bibinfo{author}{D'Amico, S.}, \bibinfo{author}{Benn, M.},  \&
  \bibinfo{author}{J{\o}rgensen, J.~L.} (\bibinfo{year}{2014}).
\newblock \bibinfo{title}{Pose estimation of an uncooperative spacecraft from
  actual space imagery}.
\newblock {\it \bibinfo{journal}{International Journal of Space Science and
  Engineering}\/},  {\it \bibinfo{volume}{2}\/}\bibinfo{issue}{(2)},
  \bibinfo{pages}{171}. \DOIprefix\doi{10.1504/ijspacese.2014.060600}.

\bibitem[{D'Amico et~al.(2013)D'Amico, Bodin, Delpech \&
  Noteborn}]{PRISMA_chapter}
\bibinfo{author}{D'Amico, S.}, \bibinfo{author}{Bodin, P.},
  \bibinfo{author}{Delpech, M.} et~al. (\bibinfo{year}{2013}).
\newblock \bibinfo{title}{{PRISMA}}.
\newblock In \bibinfo{editor}{M.~D'Errico} (Ed.), {\it
  \bibinfo{booktitle}{Distributed Space Missions for Earth System Monitoring
  Space Technology Library}\/} chapter~\bibinfo{chapter}{21}. (pp.
  \bibinfo{pages}{599--637}).
\newblock volume~\bibinfo{volume}{31}.
\newblock \DOIprefix\doi{10.1007/978-1-4614-4541-8_21}.

\bibitem[{Forshaw et~al.(2016)Forshaw, Aglietti, Navarathinam, Kadhem, Salmon,
  Pisseloup, Joffre, Chabot, Retat, Axthelm, Barraclough, Ratcliffe, Bernal,
  Chaumette, Pollini \& Steyn}]{Forshaw2016Removedebris}
\bibinfo{author}{Forshaw, J.~L.}, \bibinfo{author}{Aglietti, G.~S.},
  \bibinfo{author}{Navarathinam, N.} et~al. (\bibinfo{year}{2016}).
\newblock \bibinfo{title}{{RemoveDEBRIS}: An in-orbit active debris removal
  demonstration mission}.
\newblock {\it \bibinfo{journal}{Acta Astronautica}\/},  {\it
  \bibinfo{volume}{127}\/}, \bibinfo{pages}{448–463}.
  \DOIprefix\doi{10.1016/j.actaastro.2016.06.018}.

\bibitem[{Ganin \& Lempitsky(2015)}]{Ganin2015DANN_ICML}
\bibinfo{author}{Ganin, Y.},  \& \bibinfo{author}{Lempitsky, V.}
  (\bibinfo{year}{2015}).
\newblock \bibinfo{title}{Unsupervised domain adaptation by backpropagation}.
\newblock In \bibinfo{editor}{F.~Bach}, \& \bibinfo{editor}{D.~Blei} (Eds.),
  {\it \bibinfo{booktitle}{Proceedings of the 32nd International Conference on
  Machine Learning}\/} (pp. \bibinfo{pages}{1180--1189}).
\newblock \bibinfo{address}{Lille, France}: \bibinfo{publisher}{PMLR}
  volume~\bibinfo{volume}{37} of {\it \bibinfo{series}{Proceedings of Machine
  Learning Research}\/}.
\newblock \URLprefix \url{https://proceedings.mlr.press/v37/ganin15.html}.

\bibitem[{Ghiasi et~al.(2017)Ghiasi, Lee, Kudlur, Dumoulin \&
  Shlens}]{Ghiasi2017StyleTransfer}
\bibinfo{author}{Ghiasi, G.}, \bibinfo{author}{Lee, H.},
  \bibinfo{author}{Kudlur, M.} et~al. (\bibinfo{year}{2017}).
\newblock \bibinfo{title}{Exploring the structure of a real-time, arbitrary
  neural artistic stylization network}.
\newblock In \bibinfo{editor}{T.-K. Kim}, \bibinfo{editor}{S.~Zafeiriou},
  \bibinfo{editor}{G.~Brostow}, \& \bibinfo{editor}{K.~Mikolajczyk} (Eds.),
  {\it \bibinfo{booktitle}{Proceedings of the British Machine Vision Conference
  (BMVC)}\/} (pp. \bibinfo{pages}{114.1--114.12}).
\newblock \bibinfo{publisher}{BMVA Press}.
\newblock \DOIprefix\doi{10.5244/C.31.114}.

\bibitem[{Gidaris et~al.(2018)Gidaris, Singh \&
  Komodakis}]{Gidaris2018RotationPrediction}
\bibinfo{author}{Gidaris, S.}, \bibinfo{author}{Singh, P.},  \&
  \bibinfo{author}{Komodakis, N.} (\bibinfo{year}{2018}).
\newblock \bibinfo{title}{Unsupervised representation learning by predicting
  image rotations}.
\newblock In {\it \bibinfo{booktitle}{International Conference on Learning
  Representations}\/}.
\newblock \URLprefix \url{https://openreview.net/forum?id=S1v4N2l0-}.

\bibitem[{Ioffe \& Szegedy(2015)}]{Ioffe2015BatchNorm}
\bibinfo{author}{Ioffe, S.},  \& \bibinfo{author}{Szegedy, C.}
  (\bibinfo{year}{2015}).
\newblock \bibinfo{title}{Batch normalization: Accelerating deep network
  training by reducing internal covariate shift}.
\newblock In \bibinfo{editor}{F.~Bach}, \& \bibinfo{editor}{D.~Blei} (Eds.),
  {\it \bibinfo{booktitle}{Proceedings of the 32nd International Conference on
  Machine Learning}\/} (pp. \bibinfo{pages}{448--456}).
\newblock \bibinfo{address}{Lille, France}: \bibinfo{publisher}{PMLR}
  volume~\bibinfo{volume}{37} of {\it \bibinfo{series}{Proceedings of Machine
  Learning Research}\/}.
\newblock \URLprefix \url{https://proceedings.mlr.press/v37/ioffe15.html}.

\bibitem[{Jackson et~al.(2019)Jackson, Atapour-Abarghouei, Bonner, Breckon \&
  Obara}]{Jackson2019ICCV_StyleAug}
\bibinfo{author}{Jackson, P.~T.}, \bibinfo{author}{Atapour-Abarghouei, A.},
  \bibinfo{author}{Bonner, S.} et~al. (\bibinfo{year}{2019}).
\newblock \bibinfo{title}{Style augmentation: Data augmentation via style
  randomization}.
\newblock In {\it \bibinfo{booktitle}{Proceedings of the IEEE/CVF Conference on
  Computer Vision and Pattern Recognition (CVPR) Workshops}\/}.

\bibitem[{Kisantal et~al.(2020)Kisantal, Sharma, Park, Izzo, M\"{a}rtens \&
  D'Amico}]{Kisantal2020SPEC}
\bibinfo{author}{Kisantal, M.}, \bibinfo{author}{Sharma, S.},
  \bibinfo{author}{Park, T.~H.} et~al. (\bibinfo{year}{2020}).
\newblock \bibinfo{title}{Satellite pose estimation challenge: Dataset,
  competition design and results}.
\newblock {\it \bibinfo{journal}{IEEE Transactions on Aerospace and Electronic
  Systems}\/},  {\it \bibinfo{volume}{56}\/}\bibinfo{issue}{(5)},
  \bibinfo{pages}{4083--4098}. \DOIprefix\doi{10.1109/TAES.2020.2989063}.

\bibitem[{Knuth(1997)}]{Knuth1997ArtofComputerProgramming}
\bibinfo{author}{Knuth, D.~E.} (\bibinfo{year}{1997}).
\newblock {\it \bibinfo{title}{The art of computer programming, Volume {I:}
  Fundamental Algorithms, 3rd Edition}\/}.
\newblock \bibinfo{address}{Reading, Mass.}:
  \bibinfo{publisher}{Addison-Wesley}.

\bibitem[{Krizhevsky et~al.(2012)Krizhevsky, Sutskever \&
  Hinton}]{Krizhevsky2012AlexNet}
\bibinfo{author}{Krizhevsky, A.}, \bibinfo{author}{Sutskever, I.},  \&
  \bibinfo{author}{Hinton, G.~E.} (\bibinfo{year}{2012}).
\newblock \bibinfo{title}{Imagenet classification with deep convolutional
  neural networks}.
\newblock In \bibinfo{editor}{F.~Pereira}, \bibinfo{editor}{C.~Burges},
  \bibinfo{editor}{L.~Bottou}, \& \bibinfo{editor}{K.~Weinberger} (Eds.), {\it
  \bibinfo{booktitle}{Advances in Neural Information Processing Systems}\/}.
\newblock \bibinfo{publisher}{Curran Associates, Inc.}
  volume~\bibinfo{volume}{25}.
\newblock \URLprefix
  \url{https://proceedings.neurips.cc/paper/2012/file/c399862d3b9d6b76c8436e924a68c45b-Paper.pdf}.

\bibitem[{Lepetit et~al.(2008)Lepetit, Moreno-Noguer \& Fua}]{Lepetit2008EPnP}
\bibinfo{author}{Lepetit, V.}, \bibinfo{author}{Moreno-Noguer, F.},  \&
  \bibinfo{author}{Fua, P.} (\bibinfo{year}{2008}).
\newblock \bibinfo{title}{{EPnP}: An accurate {O(n)} solution to the {PnP}
  problem}.
\newblock {\it \bibinfo{journal}{International Journal of Computer Vision}\/},
  {\it \bibinfo{volume}{81}\/}\bibinfo{issue}{(2)}, \bibinfo{pages}{155–166}.
  \DOIprefix\doi{10.1007/s11263-008-0152-6}.

\bibitem[{Li et~al.(2020)Li, Jiao, Cao, Wong \&
  Wu}]{Li2020ECCVModelAdaptationUDA}
\bibinfo{author}{Li, R.}, \bibinfo{author}{Jiao, Q.}, \bibinfo{author}{Cao, W.}
  et~al. (\bibinfo{year}{2020}).
\newblock \bibinfo{title}{Model adaptation: Unsupervised domain adaptation
  without source data}.
\newblock In {\it \bibinfo{booktitle}{2020 IEEE/CVF Conference on Computer
  Vision and Pattern Recognition (CVPR)}\/} (pp. \bibinfo{pages}{9638--9647}).
\newblock \DOIprefix\doi{10.1109/CVPR42600.2020.00966}.

\bibitem[{Li et~al.(2017)Li, Wang, Shi, Liu \& Hou}]{Li2016AdaBN}
\bibinfo{author}{Li, Y.}, \bibinfo{author}{Wang, N.}, \bibinfo{author}{Shi, J.}
  et~al. (\bibinfo{year}{2017}).
\newblock \bibinfo{title}{Revisiting batch normalization for practical domain
  adaptation}.
\newblock \URLprefix \url{https://openreview.net/forum?id=BJuysoFeg}.

\bibitem[{Liang et~al.(2020)Liang, Hu \& Feng}]{Liang2020ICMLSHOT}
\bibinfo{author}{Liang, J.}, \bibinfo{author}{Hu, D.},  \&
  \bibinfo{author}{Feng, J.} (\bibinfo{year}{2020}).
\newblock \bibinfo{title}{Do we really need to access the source data? {S}ource
  hypothesis transfer for unsupervised domain adaptation}.
\newblock In \bibinfo{editor}{H.~D. III}, \& \bibinfo{editor}{A.~Singh} (Eds.),
  {\it \bibinfo{booktitle}{Proceedings of the 37th International Conference on
  Machine Learning}\/} (pp. \bibinfo{pages}{6028--6039}).
\newblock \bibinfo{publisher}{PMLR} volume \bibinfo{volume}{119} of {\it
  \bibinfo{series}{Proceedings of Machine Learning Research}\/}.
\newblock \URLprefix \url{https://proceedings.mlr.press/v119/liang20a.html}.

\bibitem[{Lin et~al.(2017)Lin, Goyal, Girshick, He \&
  Doll\'{a}r}]{Lin2017FocalLoss}
\bibinfo{author}{Lin, T.-Y.}, \bibinfo{author}{Goyal, P.},
  \bibinfo{author}{Girshick, R.} et~al. (\bibinfo{year}{2017}).
\newblock \bibinfo{title}{Focal loss for dense object detection}.
\newblock In {\it \bibinfo{booktitle}{2017 IEEE International Conference on
  Computer Vision (ICCV)}\/} (pp. \bibinfo{pages}{2999--3007}).
\newblock \DOIprefix\doi{10.1109/ICCV.2017.324}.

\bibitem[{Liu et~al.(2021)Liu, Kothari, van Delft, Bellot-Gurlet, Mordan \&
  Alahi}]{Liu2021TTT++}
\bibinfo{author}{Liu, Y.}, \bibinfo{author}{Kothari, P.}, \bibinfo{author}{van
  Delft, B.} et~al. (\bibinfo{year}{2021}).
\newblock \bibinfo{title}{Ttt++: When does self-supervised test-time training
  fail or thrive?}
\newblock In \bibinfo{editor}{M.~Ranzato}, \bibinfo{editor}{A.~Beygelzimer},
  \bibinfo{editor}{Y.~Dauphin}, \bibinfo{editor}{P.~Liang}, \&
  \bibinfo{editor}{J.~W. Vaughan} (Eds.), {\it \bibinfo{booktitle}{Advances in
  Neural Information Processing Systems}\/} (pp.
  \bibinfo{pages}{21808--21820}).
\newblock \bibinfo{publisher}{Curran Associates, Inc.}
  volume~\bibinfo{volume}{34}.
\newblock \URLprefix
  \url{https://proceedings.neurips.cc/paper/2021/file/b618c3210e934362ac261db280128c22-Paper.pdf}.

\bibitem[{Loshchilov \& Hutter(2019)}]{Loshchilov2017AdamW}
\bibinfo{author}{Loshchilov, I.},  \& \bibinfo{author}{Hutter, F.}
  (\bibinfo{year}{2019}).
\newblock \bibinfo{title}{Decoupled weight decay regularization}.
\newblock In {\it \bibinfo{booktitle}{International Conference on Learning
  Representations}\/}.
\newblock \URLprefix \url{https://openreview.net/forum?id=Bkg6RiCqY7}.

\bibitem[{Nath~Kundu et~al.(2020)Nath~Kundu, Venkat, Rahul \&
  Venkatesh~Babu}]{Kundu2020CVPRUniversalSourceFree}
\bibinfo{author}{Nath~Kundu, J.}, \bibinfo{author}{Venkat, N.},
  \bibinfo{author}{Rahul, M.~V.} et~al. (\bibinfo{year}{2020}).
\newblock \bibinfo{title}{Universal source-free domain adaptation}.
\newblock In {\it \bibinfo{booktitle}{2020 IEEE/CVF Conference on Computer
  Vision and Pattern Recognition (CVPR)}\/} (pp. \bibinfo{pages}{4543--4552}).
\newblock \DOIprefix\doi{10.1109/CVPR42600.2020.00460}.

\bibitem[{Noroozi \& Favaro(2016)}]{Noroozi2016ECCVJigsawPuzzle}
\bibinfo{author}{Noroozi, M.},  \& \bibinfo{author}{Favaro, P.}
  (\bibinfo{year}{2016}).
\newblock \bibinfo{title}{Unsupervised learning of visual representations by
  solving jigsaw puzzles}.
\newblock In \bibinfo{editor}{B.~Leibe}, \bibinfo{editor}{J.~Matas},
  \bibinfo{editor}{N.~Sebe}, \& \bibinfo{editor}{M.~Welling} (Eds.), {\it
  \bibinfo{booktitle}{Computer Vision -- ECCV 2016}\/} (pp.
  \bibinfo{pages}{69--84}).
\newblock \bibinfo{address}{Cham}: \bibinfo{publisher}{Springer International
  Publishing}.
\newblock \DOIprefix\doi{10.1007/978-3-319-46466-4_5}.

\bibitem[{Park et~al.(2021{\natexlab{a}})Park, Bosse \& D'Amico}]{Park2021AAS}
\bibinfo{author}{Park, T.~H.}, \bibinfo{author}{Bosse, J.},  \&
  \bibinfo{author}{D'Amico, S.} (\bibinfo{year}{2021}{\natexlab{a}}).
\newblock \bibinfo{title}{Robotic testbed for rendezvous and optical
  navigation: Multi-source calibration and machine learning use cases}.
\newblock In {\it \bibinfo{booktitle}{2021 AAS/AIAA Astrodynamics Specialist
  Conference, Big Sky, Vitrual}\/}.

\bibitem[{Park et~al.(2023)Park, M\"{a}rtens, Jawaid, Wang, Chen, Chin, Izzo \&
  D’Amico}]{park2023spec2021}
\bibinfo{author}{Park, T.~H.}, \bibinfo{author}{M\"{a}rtens, M.},
  \bibinfo{author}{Jawaid, M.} et~al. (\bibinfo{year}{2023}).
\newblock \bibinfo{title}{Satellite pose estimation competition 2021: Results
  and analyses}.
\newblock {\it \bibinfo{journal}{Acta Astronautica}\/},  {\it
  \bibinfo{volume}{204}\/}, \bibinfo{pages}{640--665}.
  \DOIprefix\doi{10.1016/j.actaastro.2023.01.002}.

\bibitem[{Park et~al.(2021{\natexlab{b}})Park, M{\"a}rtens, Lecuyer, Izzo \&
  D'Amico}]{Park2021speedplusSDR}
\bibinfo{author}{Park, T.~H.}, \bibinfo{author}{M{\"a}rtens, M.},
  \bibinfo{author}{Lecuyer, G.} et~al. (\bibinfo{year}{2021}{\natexlab{b}}).
\newblock \bibinfo{title}{Next generation spacecraft pose estimation dataset
  ({SPEED}+)}.
\newblock \bibinfo{howpublished}{Stanford Digital Repository}.
\newblock \DOIprefix\doi{10.25740/wv398fc4383} \bibinfo{note}{available at
  \url{https://purl.stanford.edu/wv398fc4383}}.

\bibitem[{Park et~al.(2022)Park, M{\"a}rtens, Lecuyer, Izzo \&
  D'Amico}]{Park2021speedplus}
\bibinfo{author}{Park, T.~H.}, \bibinfo{author}{M{\"a}rtens, M.},
  \bibinfo{author}{Lecuyer, G.} et~al. (\bibinfo{year}{2022}).
\newblock \bibinfo{title}{{SPEED+}: Next-generation dataset for spacecraft pose
  estimation across domain gap}.
\newblock In {\it \bibinfo{booktitle}{2022 IEEE Aerospace Conference (AERO)}\/}
  (pp. \bibinfo{pages}{1--15}).
\newblock \DOIprefix\doi{10.1109/AERO53065.2022.9843439}.

\bibitem[{Park et~al.(2019)Park, Sharma \& D'Amico}]{Park2019AAS}
\bibinfo{author}{Park, T.~H.}, \bibinfo{author}{Sharma, S.},  \&
  \bibinfo{author}{D'Amico, S.} (\bibinfo{year}{2019}).
\newblock \bibinfo{title}{Towards robust learning-based pose estimation of
  noncooperative spacecraft}.
\newblock In {\it \bibinfo{booktitle}{2019 AAS/AIAA Astrodynamics Specialist
  Conference, Portland, Maine}\/}.

\bibitem[{Pasqualetto~Cassinis et~al.(2019)Pasqualetto~Cassinis, Fonod \&
  Gill}]{PasqualettoCassinis2019Survey}
\bibinfo{author}{Pasqualetto~Cassinis, L.}, \bibinfo{author}{Fonod, R.},  \&
  \bibinfo{author}{Gill, E.} (\bibinfo{year}{2019}).
\newblock \bibinfo{title}{Review of the robustness and applicability of
  monocular pose estimation systems for relative navigation with an
  uncooperative spacecraft}.
\newblock {\it \bibinfo{journal}{Progress in Aerospace Sciences}\/},  {\it
  \bibinfo{volume}{110}\/}, \bibinfo{pages}{100548}.
  \DOIprefix\doi{10.1016/j.paerosci.2019.05.008}.

\bibitem[{{Pasqualetto Cassinis} et~al.(2021){Pasqualetto Cassinis}, Fonod,
  Gill, Ahrns \& Gil-Fern\'andez}]{PasqualettoCassinis2021Coupled}
\bibinfo{author}{{Pasqualetto Cassinis}, L.}, \bibinfo{author}{Fonod, R.},
  \bibinfo{author}{Gill, E.} et~al. (\bibinfo{year}{2021}).
\newblock \bibinfo{title}{Evaluation of tightly- and loosely-coupled approaches
  in cnn-based pose estimation systems for uncooperative spacecraft}.
\newblock {\it \bibinfo{journal}{Acta Astronautica}\/},  {\it
  \bibinfo{volume}{182}\/}, \bibinfo{pages}{189--202}.
  \DOIprefix\doi{10.1016/j.actaastro.2021.01.035}.

\bibitem[{{Pasqualetto Cassinis} et~al.(2022){Pasqualetto Cassinis}, Menicucci,
  Gill, Ahrns \& Sanchez-Gestido}]{PasqualettoCassinis2021ORGL}
\bibinfo{author}{{Pasqualetto Cassinis}, L.}, \bibinfo{author}{Menicucci, A.},
  \bibinfo{author}{Gill, E.} et~al. (\bibinfo{year}{2022}).
\newblock \bibinfo{title}{On-ground validation of a cnn-based monocular pose
  estimation system for uncooperative spacecraft: Bridging domain shift in
  rendezvous scenarios}.
\newblock {\it \bibinfo{journal}{Acta Astronautica}\/},  {\it
  \bibinfo{volume}{196}\/}, \bibinfo{pages}{123--138}.
  \DOIprefix\doi{10.1016/j.actaastro.2022.04.002}.

\bibitem[{Peng et~al.(2018)Peng, Usman, Kaushik, Wang, Hoffman \&
  Saenko}]{Peng2017VisDA}
\bibinfo{author}{Peng, X.}, \bibinfo{author}{Usman, B.},
  \bibinfo{author}{Kaushik, N.} et~al. (\bibinfo{year}{2018}).
\newblock \bibinfo{title}{Visda: A synthetic-to-real benchmark for visual
  domain adaptation}.
\newblock In {\it \bibinfo{booktitle}{2018 IEEE/CVF Conference on Computer
  Vision and Pattern Recognition Workshops (CVPRW)}\/} (pp.
  \bibinfo{pages}{2102--2105}).
\newblock \DOIprefix\doi{10.1109/CVPRW.2018.00271}.

\bibitem[{Proen\c{c}a \& Gao(2020)}]{Proenca2019Photorealistic}
\bibinfo{author}{Proen\c{c}a, P.~F.},  \& \bibinfo{author}{Gao, Y.}
  (\bibinfo{year}{2020}).
\newblock \bibinfo{title}{Deep learning for spacecraft pose estimation from
  photorealistic rendering}.
\newblock In {\it \bibinfo{booktitle}{2020 IEEE International Conference on
  Robotics and Automation (ICRA)}\/} (pp. \bibinfo{pages}{6007--6013}).
\newblock \DOIprefix\doi{10.1109/ICRA40945.2020.9197244}.

\bibitem[{Reed et~al.(2016)Reed, Smith, Naasz, Pellegrino \&
  Bacon}]{Reed2016RestoreL}
\bibinfo{author}{Reed, B.~B.}, \bibinfo{author}{Smith, R.~C.},
  \bibinfo{author}{Naasz, B.~J.} et~al. (\bibinfo{year}{2016}).
\newblock \bibinfo{title}{The {Restore-L} servicing mission}.
\newblock {\it \bibinfo{journal}{AIAA Space 2016}\/}, .
  \DOIprefix\doi{10.2514/6.2016-5478}.

\bibitem[{Shannon(1948)}]{Shannon1948}
\bibinfo{author}{Shannon, C.~E.} (\bibinfo{year}{1948}).
\newblock \bibinfo{title}{A mathematical theory of communication}.
\newblock {\it \bibinfo{journal}{The Bell System Technical Journal}\/},  {\it
  \bibinfo{volume}{27}\/}\bibinfo{issue}{(3)}, \bibinfo{pages}{379--423}.
  \DOIprefix\doi{10.1002/j.1538-7305.1948.tb01338.x}.

\bibitem[{Sharma \& D'Amico(2019)}]{Sharma2019AAS}
\bibinfo{author}{Sharma, S.},  \& \bibinfo{author}{D'Amico, S.}
  (\bibinfo{year}{2019}).
\newblock \bibinfo{title}{Pose estimation for non-cooperative spacecraft
  rendezvous using neural networks}.
\newblock In {\it \bibinfo{booktitle}{2019 AAS/AIAA Space Flight Mechanics
  Meeting, Ka'anapali, Maui, HI}\/}.

\bibitem[{Sharma \& D’Amico(2020)}]{Sharma2020TAES}
\bibinfo{author}{Sharma, S.},  \& \bibinfo{author}{D’Amico, S.}
  (\bibinfo{year}{2020}).
\newblock \bibinfo{title}{Neural network-based pose estimation for
  noncooperative spacecraft rendezvous}.
\newblock {\it \bibinfo{journal}{IEEE Transactions on Aerospace and Electronic
  Systems}\/},  {\it \bibinfo{volume}{56}\/}\bibinfo{issue}{(6)},
  \bibinfo{pages}{4638--4658}. \DOIprefix\doi{10.1109/TAES.2020.2999148}.

\bibitem[{Sharma et~al.(2019)Sharma, Park \& D'Amico}]{Sharma2019SPEEDonSDR}
\bibinfo{author}{Sharma, S.}, \bibinfo{author}{Park, T.~H.},  \&
  \bibinfo{author}{D'Amico, S.} (\bibinfo{year}{2019}).
\newblock \bibinfo{title}{Spacecraft pose estimation dataset ({SPEED})}.
\newblock \bibinfo{howpublished}{Stanford Digital Repository.}
\newblock \DOIprefix\doi{10.25740/dz692fn7184} \bibinfo{note}{available at:
  \url{https://purl.stanford.edu/dz692fn7184}}.

\bibitem[{Shorten \& Khoshgoftaar(2019)}]{Shorten2019DataAugmentation}
\bibinfo{author}{Shorten, C.},  \& \bibinfo{author}{Khoshgoftaar, T.~M.}
  (\bibinfo{year}{2019}).
\newblock \bibinfo{title}{A survey on image data augmentation for deep
  learning}.
\newblock {\it \bibinfo{journal}{Journal of Big Data}\/},  {\it
  \bibinfo{volume}{6}\/}. \DOIprefix\doi{10.1186/s40537-019-0197-0}.

\bibitem[{Sun \& Saenko(2016)}]{Sun2016ECCV_DeepCORAL}
\bibinfo{author}{Sun, B.},  \& \bibinfo{author}{Saenko, K.}
  (\bibinfo{year}{2016}).
\newblock \bibinfo{title}{Deep coral: Correlation alignment for deep domain
  adaptation}.
\newblock In \bibinfo{editor}{G.~Hua}, \& \bibinfo{editor}{H.~J{\'e}gou}
  (Eds.), {\it \bibinfo{booktitle}{Computer Vision -- ECCV 2016 Workshops}\/}
  (pp. \bibinfo{pages}{443--450}).
\newblock \bibinfo{address}{Cham}: \bibinfo{publisher}{Springer International
  Publishing}.
\newblock \DOIprefix\doi{10.1007/978-3-319-49409-8_35}.

\bibitem[{Sun et~al.(2020)Sun, Wang, Liu, Miller, Efros \&
  Hardt}]{Sun19TestTimeTraining}
\bibinfo{author}{Sun, Y.}, \bibinfo{author}{Wang, X.}, \bibinfo{author}{Liu,
  Z.} et~al. (\bibinfo{year}{2020}).
\newblock \bibinfo{title}{Test-time training with self-supervision for
  generalization under distribution shifts}.
\newblock In \bibinfo{editor}{H.~D. III}, \& \bibinfo{editor}{A.~Singh} (Eds.),
  {\it \bibinfo{booktitle}{Proceedings of the 37th International Conference on
  Machine Learning}\/} (pp. \bibinfo{pages}{9229--9248}).
\newblock \bibinfo{publisher}{PMLR} volume \bibinfo{volume}{119} of {\it
  \bibinfo{series}{Proceedings of Machine Learning Research}\/}.
\newblock \URLprefix \url{https://proceedings.mlr.press/v119/sun20b.html}.

\bibitem[{Tan \& Le(2019)}]{Tan2019EfficientNetICML}
\bibinfo{author}{Tan, M.},  \& \bibinfo{author}{Le, Q.} (\bibinfo{year}{2019}).
\newblock \bibinfo{title}{{E}fficient{N}et: Rethinking model scaling for
  convolutional neural networks}.
\newblock In \bibinfo{editor}{K.~Chaudhuri}, \&
  \bibinfo{editor}{R.~Salakhutdinov} (Eds.), {\it
  \bibinfo{booktitle}{Proceedings of the 36th International Conference on
  Machine Learning}\/} (pp. \bibinfo{pages}{6105--6114}).
\newblock \bibinfo{publisher}{PMLR} volume~\bibinfo{volume}{97} of {\it
  \bibinfo{series}{Proceedings of Machine Learning Research}\/}.
\newblock \URLprefix \url{https://proceedings.mlr.press/v97/tan19a.html}.

\bibitem[{Tan et~al.(2020)Tan, Pang \& Le}]{Tan2020EfficientDetCVPR}
\bibinfo{author}{Tan, M.}, \bibinfo{author}{Pang, R.},  \& \bibinfo{author}{Le,
  Q.~V.} (\bibinfo{year}{2020}).
\newblock \bibinfo{title}{{E}fficient{D}et: Scalable and efficient object
  detection}.
\newblock In {\it \bibinfo{booktitle}{2020 IEEE/CVF Conference on Computer
  Vision and Pattern Recognition (CVPR)}\/} (pp.
  \bibinfo{pages}{10778--10787}).
\newblock \DOIprefix\doi{10.1109/CVPR42600.2020.01079}.

\bibitem[{Tobin et~al.(2017)Tobin, Fong, Ray, Schneider, Zaremba \&
  Abbeel}]{Tobin2017DomainRandomization}
\bibinfo{author}{Tobin, J.}, \bibinfo{author}{Fong, R.}, \bibinfo{author}{Ray,
  A.} et~al. (\bibinfo{year}{2017}).
\newblock \bibinfo{title}{Domain randomization for transferring deep neural
  networks from simulation to the real world}.
\newblock In {\it \bibinfo{booktitle}{2017 IEEE/RSJ International Conference on
  Intelligent Robots and Systems (IROS)}\/} (pp. \bibinfo{pages}{23--30}).
\newblock \DOIprefix\doi{10.1109/IROS.2017.8202133}.

\bibitem[{Tzeng et~al.(2017)Tzeng, Hoffman, Saenko \&
  Darrell}]{Tzeng2017CVPR_ADDA}
\bibinfo{author}{Tzeng, E.}, \bibinfo{author}{Hoffman, J.},
  \bibinfo{author}{Saenko, K.} et~al. (\bibinfo{year}{2017}).
\newblock \bibinfo{title}{Adversarial discriminative domain adaptation}.
\newblock In {\it \bibinfo{booktitle}{2017 IEEE Conference on Computer Vision
  and Pattern Recognition (CVPR)}\/} (pp. \bibinfo{pages}{2962--2971}).
\newblock \DOIprefix\doi{10.1109/CVPR.2017.316}.

\bibitem[{Wang et~al.(2021)Wang, Shelhamer, Liu, Olshausen \&
  Darrell}]{Wang2021ICLRTENT}
\bibinfo{author}{Wang, D.}, \bibinfo{author}{Shelhamer, E.},
  \bibinfo{author}{Liu, S.} et~al. (\bibinfo{year}{2021}).
\newblock \bibinfo{title}{Tent: Fully test-time adaptation by entropy
  minimization}.
\newblock In {\it \bibinfo{booktitle}{International Conference on Learning
  Representations}\/}.
\newblock \URLprefix \url{https://openreview.net/forum?id=uXl3bZLkr3c}.

\bibitem[{Wu \& He(2018)}]{Wu2018ECCV_GroupNorm}
\bibinfo{author}{Wu, Y.},  \& \bibinfo{author}{He, K.} (\bibinfo{year}{2018}).
\newblock \bibinfo{title}{Group normalization}.
\newblock In \bibinfo{editor}{V.~Ferrari}, \bibinfo{editor}{M.~Hebert},
  \bibinfo{editor}{C.~Sminchisescu}, \& \bibinfo{editor}{Y.~Weiss} (Eds.), {\it
  \bibinfo{booktitle}{Computer Vision -- ECCV 2018}\/} (pp.
  \bibinfo{pages}{3--19}).
\newblock \bibinfo{address}{Cham}: \bibinfo{publisher}{Springer International
  Publishing}.
\newblock \DOIprefix\doi{10.1007/978-3-030-01261-8_1}.

\bibitem[{Zheng et~al.(2020)Zheng, Wang, Liu, Li, Ye \& Ren}]{Zheng2020DIoU}
\bibinfo{author}{Zheng, Z.}, \bibinfo{author}{Wang, P.}, \bibinfo{author}{Liu,
  W.} et~al. (\bibinfo{year}{2020}).
\newblock \bibinfo{title}{{Distance-IoU} loss: Faster and better learning for
  bounding box regression}.
\newblock {\it \bibinfo{journal}{Proceedings of the AAAI Conference on
  Artificial Intelligence}\/},  {\it
  \bibinfo{volume}{34}\/}\bibinfo{issue}{(07)}, \bibinfo{pages}{12993--13000}.
  \DOIprefix\doi{10.1609/aaai.v34i07.6999}.

\bibitem[{Zhong et~al.(2020)Zhong, Zheng, Kang, Li \&
  Yang}]{Zhong2020RandomErasing}
\bibinfo{author}{Zhong, Z.}, \bibinfo{author}{Zheng, L.},
  \bibinfo{author}{Kang, G.} et~al. (\bibinfo{year}{2020}).
\newblock \bibinfo{title}{Random erasing data augmentation}.
\newblock {\it \bibinfo{journal}{Proceedings of the AAAI Conference on
  Artificial Intelligence}\/},  {\it
  \bibinfo{volume}{34}\/}\bibinfo{issue}{(07)}, \bibinfo{pages}{13001--13008}.
  \DOIprefix\doi{10.1609/aaai.v34i07.7000}.

\bibitem[{Zhou et~al.(2019)Zhou, Barnes, Lu, Yang \&
  Li}]{Zhou2019RotationParametrization}
\bibinfo{author}{Zhou, Y.}, \bibinfo{author}{Barnes, C.}, \bibinfo{author}{Lu,
  J.} et~al. (\bibinfo{year}{2019}).
\newblock \bibinfo{title}{On the continuity of rotation representations in
  neural networks}.
\newblock In {\it \bibinfo{booktitle}{2019 IEEE/CVF Conference on Computer
  Vision and Pattern Recognition (CVPR)}\/} (pp. \bibinfo{pages}{5738--5746}).
\newblock \DOIprefix\doi{10.1109/CVPR.2019.00589}.

\end{thebibliography}

\end{document}